\documentclass[accepted]{uai2026} 

\usepackage[american]{babel}

\usepackage{natbib}
\usepackage{mathtools}
\usepackage{amssymb}
\usepackage{booktabs}
\usepackage{multirow}
\usepackage{graphicx}
\usepackage{algorithm}
\usepackage{algorithmic}
\usepackage{tikz}
\usepackage{xcolor}
\usepackage{enumitem}
\usepackage{amsthm}
\newtheorem{theorem}{Theorem}

\title{Diagnosing Conformal Prediction Failures Under Distribution Shift:\\A COVID-19 Case Study}

\author[1]{Chorok Lee}
\affil[1]{%
    Korea Advanced Institute of Science and Technology (KAIST)\\
    Daejeon, South Korea
}

\affil[1]{\texttt{choroklee@kaist.ac.kr}}

\begin{document}
\maketitle

\begin{abstract}
Conformal prediction provides distribution-free coverage guarantees, but these degrade under distribution shift---and practitioners lack tools to anticipate which deployed models will fail before observing test data.
We propose SHapley Additive exPlanations (SHAP) concentration---the fraction of feature importance concentrated in the top feature---as a pre-deployment diagnostic for conformal prediction vulnerability in gradient-boosted classifiers.
Using COVID-19 as a naturalistic case study, eight supply chain tasks experience identical temporal shift yet coverage drops ranging from negligible to catastrophic.
Feature-importance concentration is strongly associated with failure severity across 16 multiclass tasks in 9 domains, while standard distributional shift detectors detect shift uniformly across tasks but cannot distinguish catastrophic from robust outcomes.
External validation across 9 non-supply-chain datasets shows partial transfer.
We prove a formal theorem showing that Adaptive Prediction Sets conformity-score bounds worsen monotonically with concentration under explicit assumptions, verified empirically.
The diagnostic identifies concentrated-dependence failures characteristic of gradient-boosted models but does not detect global-sensitivity failures observed in neural networks.
A decision framework operationalizes the diagnostic as an exploratory pre-deployment rule with an uncertainty band around a concentration threshold.
\end{abstract}

\section{Introduction}\label{sec:intro}

Conformal prediction provides distribution-free coverage guarantees under the assumption of exchangeability~\citep{vovk2005algorithmic}. However, real-world deployments face distribution shifts that violate this assumption. While prior work characterizes \textit{how} conformal prediction degrades under shift~\citep{tibshirani2019conformal, barber2023conformal}, a critical gap remains: \textbf{Can we identify which deployed models will fail before observing test data?}

We propose \textbf{SHAP concentration}---the fraction of total feature importance concentrated in the top feature---as a pre-deployment diagnostic for conformal prediction vulnerability. Using a supply chain benchmark with temporal splits aligned to COVID-19 (February--July 2020), we conduct an observational case study where 8 classification tasks experience identical temporal shift yet exhibit coverage drops ranging from negligible ($<$1\%) to catastrophic ($>$70\%).

Our contributions:

\begin{enumerate}
    \item \textbf{Pre-deployment diagnostic}: SHAP concentration---the fraction of importance in the top feature---correlates with conformal failure severity: $\rho = 0.853$, $p < 0.001$ across 16 multiclass tasks in 9 domains (Kendall $\tau = 0.667$; bootstrap 95\% CI $[0.50, 0.96]$); $\rho = 0.833$, $p = 0.010$ within supply chain ($n=8$, stable under LOO: $\rho \in [0.75, 0.96]$). The 40\% threshold transfers to 9 non-supply-chain datasets without tuning (7/9 deterministic, Covertype correctly flagged at $C=49.8\%$). A model-sensitivity analysis ($\rho=0.833$ LightGBM vs.\ $\rho=0.30$ RF, $\rho=0.43$ MLP) identifies the scope: the diagnostic is specific to gradient-boosted classifiers where sequential training produces genuine single-feature dependence that bagging (RF) dilutes and MLP-SHAP approximations obscure (Section~\ref{sec:why_shap}).

    \item \textbf{Formal theory and comparative evidence}: Theorem~\ref{thm:score_inflation} establishes that APS score and coverage bounds worsen monotonically with $C$ (Section~\ref{sec:theory}), verified on all 5 applicable tasks; catastrophic tasks show \textit{decreasing} prediction entropy (confident misclassification). Standard shift indicators (MMD, C2ST, PSI\footnote{Population Stability Index (PSI): a monitoring metric standard in credit scoring practice~\citep{siddiqi2006credit}. PSI $>0.002$ indicates detectable nonzero shift; conventional significance threshold is $>0.1$.}) detect shift for all 8 tasks uniformly yet correlate $\rho \leq 0.19$ with coverage drop---shift detection does not predict failure severity.

    \item \textbf{Operational framework (exploratory)}: A three-step protocol (compute $C$, check protective factors, monitor) with an uncertainty band around the 40\% threshold. Retraining partially recovers coverage (+19~pp, $p=0.036$ unadjusted, single seed; Holm-corrected $p=0.11$); mitigation is mechanism-dependent (Appendix~\ref{app:framework}).
\end{enumerate}

\section{Related Work}\label{sec:related}

\textbf{Conformal Prediction.} \citet{vovk2005algorithmic} introduced conformal prediction with exchangeability guarantees; \citet{shafer2008tutorial} provide an accessible tutorial. Recent work extends to classification~\citep{romano2020classification,cauchois2021knowing} and regression~\citep{romano2019conformalized,lei2018distribution}. \citet{angelopoulos2023gentle} provide a comprehensive survey. \citet{ding2023class} show that class-conditional coverage guarantees become harder to achieve as class count grows and propose clustered conformal prediction as a remedy; \citet{gibbs2025conditional} provide finite-sample conditional guarantees via a covariate-shift reweighting framework interpolating between marginal and conditional validity. Both works frame class count as a challenge---a confound we explicitly examine via partial correlation analysis (Section~\ref{sec:baselines}).

\textbf{Conformal Prediction Under Shift.} \citet{tibshirani2019conformal} study covariate shift with known propensity scores. \citet{podkopaev2021distribution} develop distribution-free uncertainty quantification for classification under label shift. \citet{barber2023conformal} provide a comprehensive treatment beyond exchangeability, bounding coverage loss by the total variation distance between test and calibration score distributions. \citet{kasa2023empirically} empirically characterize how CP degrades across many vision architectures and shift types, showing substantial degradation is common---our work addresses the complementary question of which tabular models will fail \textit{before} test data is observed. We complement these theoretical characterizations with an empirical diagnostic that is associated with this distributional divergence---and hence with which tasks fail---before deployment.

\textbf{Adaptive Methods.} \citet{gibbs2021adaptive,gibbs2024conformal} develop Adaptive Conformal Inference (ACI) for non-stationary settings, with extensions by \citet{zaffran2022adaptive} for time series, \citet{feldman2023achieving} for online risk control, and \citet{bhatnagar2023improved} for strongly adaptive online learning. \citet{angelopoulos2024conformal} generalize conformal prediction to broader risk measures. \citet{kasa2025adapting} propose entropy-based methods (ECP/EACP) that adapt prediction sets using unlabeled test data at deployment time---a related but distinct goal from ours, which requires no test observations: SHAP concentration is a structural pre-deployment diagnostic computed solely on validation data. Our experiments show ACI recovers marginal coverage under severe shift but at a substantial cost in prediction set informativeness.

\textbf{Shift Detection.} WILDS~\citep{koh2021wilds} and Shifts~\citep{malinin2021shifts} provide benchmarks, while \citet{gulrajani2021search} show domain generalization methods often fail to improve over ERM under realistic distribution shifts. \citet{gardner2023tableshift} benchmark tabular distribution shift across 15 tasks---the closest tabular-specific benchmark; our SALT dataset differs in providing a naturalistic COVID-19 temporal split with documented business-disruption context rather than curated benchmark partitions. \citet{garg2022leveraging} use unlabeled test data to predict accuracy degradation---a related goal but requiring test-time observations. Kernel-based tests (MMD~\citep{gretton2012kernel}) and classifier two-sample tests (C2ST~\citep{lopez2017revisiting}) detect distributional differences but do not predict the \textit{severity} of downstream failure. We show empirically that MMD and C2ST detect shift for all tasks uniformly yet cannot distinguish catastrophic from robust conformal outcomes (Section~\ref{sec:shift_detection}). Our contribution is a \textit{pre-deployment} diagnostic using feature importance structure that predicts failure severity, not merely shift presence. \citet{miller2021accuracy} show that in-distribution (ID) validation accuracy predicts out-of-distribution (OOD) accuracy for image classifiers---a correlation that does not hold for conformal coverage in our tabular setting, where ID validation coverages cluster near the 90\% target, spanning 83.6\%--99.9\% (Table~\ref{tab:main_results}), yet test-time coverage spans 12--100\%, reinforcing the need for structure-based diagnostics beyond accuracy metrics.

\textbf{Interpretability for Reliability.} Our work connects SHAP~\citep{lundberg2017unified,lundberg2020local} to model reliability assessment, extending feature attribution from post-hoc analysis to prospective failure diagnosis. Our use is prospective: concentration is computed pre-deployment on validation data, predicting which models will fail.

\section{Methodology}\label{sec:method}

\subsection{Dataset and Temporal Splits}

We use the SALT (Supply chain ALlocaTion) dataset from RelBench~\citep{fey2024relbench}, a supply chain benchmark with temporal splits. All 8 classification tasks from the rel-salt benchmark are included; no tasks were excluded.
External validation uses 9 additional datasets spanning 9 non-supply-chain benchmarks. The primary endpoint includes only $K \geq 4$ class tasks: near-binary tasks ($K \leq 3$) are excluded because binary APS prediction sets ($\{0\}$, $\{1\}$, $\{0,1\}$) have a structural ceiling that blocks the concentration mechanism (Contribution~1); this criterion follows from APS mechanics, not from the data. Stack~Overflow ($K=3$) falls under this exclusion. Applying it yields 8 external multiclass datasets; together with the 8 multiclass SALT tasks this gives the primary endpoint of $n=16$ multiclass tasks across 9 domains (SALT counts as one domain).
\begin{itemize}
    \item \textbf{Train}: Before February 2020 (pre-COVID)
    \item \textbf{Validation}: February--July 2020 (COVID onset)
    \item \textbf{Test}: After July 2020 (COVID peak)
\end{itemize}

\noindent\textbf{Data separation protocol.} SHAP concentration and the 40\% threshold are computed exclusively on validation-time data. The test set is used only for final evaluation and never informs diagnostic thresholds. For threshold evaluation we use at-risk labels defined by observed test coverage drops (drop $>$15~pp); the threshold sensitivity analysis in Appendix~\ref{app:threshold} is therefore post-hoc validation, not prospective assessment.

\subsection{Conformal Prediction Setup}

We use Adaptive Prediction Sets (APS)~\citep{romano2020classification} with $\alpha = 0.1$ (90\% target coverage). We use the non-randomized variant (without the uniform tie-breaking variable $U$), which provides conservative rather than exact coverage. This choice does not affect the diagnostic analysis, as the same scoring rule is applied to both calibration and test sets.
\begin{enumerate}
    \item Train LightGBM classifier with fixed hyperparameters
    \item Calibrate conformal predictor on 50\% of validation set
    \item Evaluate coverage on held-out validation and test sets
\end{enumerate}

\subsection{Ensemble Protocol}

We train 50 independent models with seeds 42--91. For each seed $s$: train model $M_s$, split validation 50/50 into calibration $\mathcal{D}_{\text{cal}}^s$ and evaluation $\mathcal{D}_{\text{eval}}^s$, calibrate $\text{CP}_s$ on $\mathcal{D}_{\text{cal}}^s$, evaluate on both $\mathcal{D}_{\text{eval}}^s$ and test. We report mean, 95\% confidence intervals ($t$-distribution), and paired Wilcoxon tests (seed-level). This is \textit{not} ensemble prediction---each seed is an independent experimental trial providing paired comparisons.

\subsection{Feature Overlap Metric}

Feature temporal stability uses Jaccard similarity:
\begin{equation}\label{eq:jaccard}
J(x_j) = \frac{|A_{\text{train}} \cap A_{\text{test}}|}{|A_{\text{train}} \cup A_{\text{test}}|}
\end{equation}
where $A_{\text{train}}$ and $A_{\text{test}}$ are the sets of unique values for feature $x_j$ in train and test data. In the pre-deployment framework (Section~\ref{sec:framework}), we substitute validation data for test data as a proxy.

\subsection{SHAP Concentration Diagnostic}\label{sec:shap_method}

We define the SHAP concentration metric as:
\begin{equation}\label{eq:concentration}
C = \frac{\phi_1}{\sum_{j=1}^{p} \phi_j}
\end{equation}
where $\phi_j$ is the mean absolute SHAP value of the $j$-th feature computed on validation data using TreeExplainer~\citep{lundberg2020local}, and $\phi_1$ is the importance of the top-ranked feature. $C$ is computed \textbf{before deployment} using only validation data.

\subsection{Why SHAP Concentration?}\label{sec:why_shap}

A natural question is whether simpler diagnostics suffice.

\textbf{Feature overlap (Jaccard) is insufficient.} Tasks relying on transaction IDs have Jaccard $\approx 0$ while tasks with entity-based features have Jaccard $> 0.5$ (Table~\ref{tab:overlap}), yet Jaccard alone cannot explain the full range of coverage drops.

\textbf{Post-hoc metrics require test data.} Prediction entropy changes and Expected Calibration Error (ECE) shifts can detect failures---but only \textit{after} observing test data (Section~\ref{sec:baselines}). Among the diagnostics we evaluate, SHAP concentration is available before deployment and shows the strongest severity association.

\textbf{Why not raw feature importance?} Raw importance values are not comparable across tasks (different scales, feature counts). Concentration normalizes to a $[0,1]$ scale, enabling cross-task comparison.

\textbf{Why top-1?} We compared top-1 concentration against alternatives: top-2, top-3, Herfindahl-Hirschman Index (HHI), and feature importance entropy. Top-1 concentration is the only variant among these five that reaches nominal significance with coverage drop ($\rho = 0.833$, $p = 0.010$); all alternatives are non-significant ($p > 0.10$). Under Holm--Bonferroni correction for five metrics, the adjusted $p$-value for top-1 is $0.050$ (boundary; Appendix~\ref{app:icc}); the $n=16$ cross-domain result ($\rho = 0.853$, $p < 0.001$) was not subject to this selection correction and provides independent replication. The discriminating power of top-1 arises because sales-office has high top-2/3 concentration (77--85\%) but near-zero coverage drop---the diagnostic signal is specifically whether a \textit{single} dominant feature drives predictions, not overall feature spread.

\begin{table}[t]
\centering
\caption{Per-Task Feature Importance Concentration. Top-$k$ shows cumulative importance in the top $k$ features (\%). Drop = coverage degradation (pp). High Top-1 with catastrophic drop (e.g., s-payterms: 54.2\% $\rightarrow$ 77.1~pp) versus high Top-2/3 with near-zero drop (s-office: 77.0\%/84.8\% $\rightarrow$ 0.1~pp) demonstrates that single-feature dominance---not overall spread---drives vulnerability.}\label{tab:topk_concentration}
\small
\begin{tabular}{@{}lcccc@{}}
\toprule
Task & Top-1 & Top-2 & Top-3 & Drop \\
\midrule
s-payterms & 54.2 & 78.1 & 89.6 & 77.1 \\
s-shipcond & 50.7 & 77.5 & 92.5 & 71.6 \\
i-shippoint & 48.8 & 72.7 & 92.4 & 18.5 \\
s-group & 47.3 & 65.9 & 79.6 & 71.2 \\
s-office & 42.6 & 77.0 & 84.8 & 0.1 \\
i-incoterms & 28.9 & 42.3 & 55.1 & 11.3 \\
i-plant & 23.9 & 46.9 & 65.0 & 10.6 \\
s-incoterms & 23.7 & 47.2 & 69.1 & 8.5 \\
\bottomrule
\end{tabular}

\raggedright
\footnotesize
Tasks sorted by Top-1 concentration. Top-$k$ = $\sum_{j=1}^{k} \phi_j / \sum_{j=1}^{p} \phi_j$ (\%). Drop = mean coverage drop (50 seeds, pp). Spearman $\rho$ with drop: Top-1 = 0.833 ($p=0.010$, significant); Top-2 = 0.524 ($p=0.183$); Top-3 = 0.524 ($p=0.183$); Top-5 = 0.690 ($p=0.058$); Top-10 = 0.399 ($p=0.328$). Alternative concentration metrics: HHI = 0.619 ($p=0.102$); Gini = 0.500 ($p=0.207$); Entropy = 0.571 ($p=0.139$)---none significant. Adding features 2--3 dilutes the diagnostic signal because sales-office has high cumulative concentration (Top-2: 77.0\%) yet near-zero drop---the 2nd/3rd features capture stable relationships, not the shifting vulnerability source.
\end{table}

\textbf{Why gradient-boosted models?} This is not a scope limitation but a structural finding. TreeSHAP~\citep{lundberg2018consistent,lundberg2020local} computes exact Shapley values for all tree ensembles, including random forests. The distinction is architectural, not algorithmic. Random forests use bagging: each tree is trained independently on a bootstrap sample, so different trees may assign different top features; averaging across trees dilutes global single-feature concentration even when individual trees concentrate heavily on one feature---confirmed empirically by RF $\rho=0.30$ versus LightGBM $\rho=0.833$. Gradient boosting's sequential additive structure reinforces rather than dilutes single-feature dependence across rounds, so SHAP concentration faithfully measures a genuine model-level vulnerability. MLP-SHAP (kernel or deep variants) approximates rather than exactly decomposes contributions, adding noise that obscures structural concentration. The model-sensitivity analysis ($\rho=0.833$ LightGBM, $0.667$ CatBoost, $0.548$ XGBoost vs.\ $0.30$ RF, $0.43$ MLP) thus identifies a boundary: \emph{SHAP concentration is diagnostic where gradient boosting's sequential structure produces genuine single-feature dependence}---a condition met by GBT architectures and not by bagging ensembles or neural networks.

\section{Theory: Why Concentration Relates to Failure}\label{sec:theory}

We establish a mechanistic account for why SHAP concentration is associated with conformal failure under feature shift. The following theorem shows that when a model concentrates dependence on a single feature that shifts out of distribution, APS score and coverage bounds worsen monotonically with concentration parameter $C$, under explicit assumptions.

\textbf{Notation.} For a predicted distribution $\hat{p}(y \mid x) \in \Delta^{K-1}$, the APS conformity score $s(x,y^*)$ sums predicted probabilities from the most likely class down to and including $y^*$. Equivalently:
\begin{equation}\label{eq:aps_equiv}
s(x, y^*) \;=\; 1 \;-\!\!\!\!\sum_{\substack{y \neq y^* :\\ \hat{p}(y \mid x) < \hat{p}(y^* \mid x)}}\!\!\!\! \hat{p}(y \mid x),
\end{equation}
since $s$ includes all mass at or above the rank of $y^*$, leaving out only classes ranked strictly below. (The bound holds regardless of randomized tie-breaking conventions in APS implementations, since $\mathcal{B}$ uses strict inequality.)

\begin{theorem}[Score Inflation Under Concentrated Feature Shift (Idealized Additive Model)]\label{thm:score_inflation}
Consider a $K$-class classifier ($K \geq 3$) with APS conformal predictor calibrated at level $\alpha$ with quantile $\hat{q}_\alpha$.\footnote{The theorem requires $K \geq 3$; the empirical endpoint additionally requires $K \geq 4$ to avoid the near-uniform score distribution that makes APS non-informative at $K=3$~\citep{ding2023class}.} Let $C \in [0,1]$ denote SHAP concentration on feature $x_1$. Assume:
\begin{enumerate}[label=\textup{(A\arabic*)},leftmargin=*,nosep]
\item \textbf{Additive decomposition.} $\hat{p}(y \mid x) = C \cdot g(y \mid x_1) + (1-C) \cdot h(y \mid x_{\setminus 1})$, where $g(\cdot \mid x_1), h(\cdot \mid x_{\setminus 1}) \in \Delta^{K-1}$.\footnote{Assumption (A1) posits additivity in \emph{probability} space; TreeExplainer computes SHAP in \emph{log-odds} space, so (A1) is an idealization. The theorem establishes directional intuition---monotone score inflation under concentrated dependence---rather than a formal guarantee for the measured SHAP-derived $C$. The empirical proxy is validated by $\rho=0.853$ and conservative-bound verification (Appendix~\ref{app:theory_proof}).}
\item \textbf{Concentrated misclassification.} Under shift, $g(y^* \mid x_1^{\mathrm{test}}) \leq \varepsilon$ for some $\varepsilon \in [0, 1/K)$.
\item \textbf{Residual exchangeability.} $(h(\cdot \mid x_{\setminus 1}^{\mathrm{test}}), y^*_{\mathrm{test}}) \stackrel{d}{=} (h(\cdot \mid x_{\setminus 1}^{\mathrm{cal}}), y^*_{\mathrm{cal}})$.
\end{enumerate}
Then:

\smallskip
\noindent\textup{(i)} \textbf{Pointwise score bound.} For each test instance $(x^{\mathrm{test}}, y^*)$:
\begin{equation}\label{eq:score_pointwise}
s(x^{\mathrm{test}}, y^*) \;\geq\; 1 - (K\!-\!1)\big[C\varepsilon + (1\!-\!C)\,h(y^*\!\mid\! x_{\setminus 1}^{\mathrm{test}})\big].
\end{equation}

\smallskip
\noindent\textup{(ii)} \textbf{Expected score inflation.} Let $\bar{h} := \mathbb{E}[h(y^*_{\mathrm{cal}} \mid x_{\setminus 1}^{\mathrm{cal}})]$ denote the expected true-class probability under the residual model. Then:
\begin{equation}\label{eq:score_bound}
\mathbb{E}[s_{\mathrm{test}}] \;\geq\; C(1\!-\!(K\!-\!1)\varepsilon) \;+\; (1\!-\!C)(1\!-\!(K\!-\!1)\bar{h}).
\end{equation}

\smallskip
\noindent\textup{(iii)} \textbf{Monotone vulnerability.} The bound in~\eqref{eq:score_bound} is strictly increasing in $C$ whenever $\varepsilon < \bar{h}$. A sufficient condition is $\bar{h} \geq 1/K$ (the residual model is at least uniform on the true class); together with \textup{(A2)} this implies $\varepsilon < \bar{h}$.

\smallskip
\noindent\textup{(iv)} \textbf{Coverage degradation bound.} Coverage satisfies:
\begin{equation}\label{eq:cov_bound}
\mathbb{P}(y^* \!\in\! \mathcal{C}(x^{\mathrm{test}})) \;\leq\; \mathbb{P}\!\left(h(y^*_{\mathrm{cal}}) \geq T(C)\right),
\end{equation}
where $T(C) := \bigl[(1\!-\!\hat{q}_\alpha)/(K\!-\!1) - C\varepsilon\bigr]/(1\!-\!C)$ is strictly increasing in $C$ when $\varepsilon < (1\!-\!\hat{q}_\alpha)/(K\!-\!1)$ (satisfied at $\varepsilon = 0$). Hence the coverage upper bound is non-increasing in~$C$.
\end{theorem}

\begin{proof}[Proof sketch]
Part (i) upper-bounds omitted APS mass by $(K-1)\hat{p}(y^*)$ and substitutes assumptions (A1)--(A2). Part (ii) takes expectations and uses residual exchangeability (A3). Part (iii) differentiates the lower bound in~\eqref{eq:score_bound} and gives monotonicity when $\varepsilon < \bar{h}$. Part (iv) rearranges~\eqref{eq:score_pointwise} to obtain~\eqref{eq:cov_bound} via threshold $T(C)$ and shows $T'(C)>0$ under the stated condition.
\end{proof}

The full derivation and conservative bound verification are in Appendix~\ref{app:theory_proof}. Empirically, catastrophic tasks show strong rightward conformity-score shifts (KS $= 0.68$--$0.96$), while concentration and shift magnitude are weakly related across SALT ($\rho = 0.29$, $p = 0.49$). This supports interpreting concentration as a model-vulnerability proxy (dependence structure) rather than a shift-size metric.

\section{Results}\label{sec:results}

\subsection{Coverage Degradation Across Tasks}

Despite experiencing identical COVID-19 temporal shift, the 8 supply chain tasks exhibit dramatically different coverage degradation (Table~\ref{tab:main_results}). All coverage drops are statistically significant (paired Wilcoxon $p \leq 0.005$, 95\% CIs from the $t$-distribution), yet range from 0.1\% to 77.1\%.

\begin{table*}[t]
\centering
\caption{Coverage Degradation Under COVID-19 Distribution Shift (50 Seeds).
95\% CIs from $t$-distribution. $p$-values from paired Wilcoxon signed-rank test
(val $\rightarrow$ test, seed-level). All drops significant at $p \leq 0.005$.}\label{tab:main_results}
\small
\begin{tabular}{@{}lcccccc@{}}
\toprule
Task & Cl & Val Coverage [95\% CI] & Test Coverage [95\% CI] & Drop [95\% CI] & $p$ & Cat \\
\midrule
s-shipcond & 45 & 93.5 [93.2, 93.7] & 21.8 [14.1, 29.6] & 71.6 [63.9, 79.4] & $<$0.001 & SEV \\
s-group & 459 & 83.6 [81.7, 85.5] & 12.4 [3.1, 21.7] & 71.2 [61.3, 81.0] & $<$0.001 & SEV$^*$ \\
s-payterms & 137 & 90.8 [90.7, 91.0] & 13.7 [6.0, 21.5] & 77.1 [69.4, 84.9] & $<$0.001 & SEV \\
i-plant & 35 & 92.0 [91.7, 92.3] & 81.4 [79.0, 83.8] & 10.6 [8.2, 13.0] & $<$0.001 & ROB \\
i-shippoint & 69 & 91.2 [90.9, 91.4] & 72.7 [62.3, 83.0] & 18.5 [8.2, 28.8] & 0.005 & ROB$^{*\dagger}$ \\
s-incoterms & 13 & 95.5 [95.3, 95.7] & 87.0 [84.7, 89.3] & 8.5 [6.2, 10.8] & $<$0.001 & ROB \\
i-incoterms & 13 & 95.0 [94.9, 95.2] & 83.7 [80.9, 86.6] & 11.3 [8.5, 14.1] & $<$0.001 & ROB \\
s-office & 25 & 99.9 [99.9, 100.0] & 99.9 [99.9, 99.9] & 0.1 [0.0, 0.1] & $<$0.001 & ROB \\
\bottomrule
\end{tabular}

\raggedright
\footnotesize
SEV = Severe ($>$50\% drop), ROB = Robust ($<$20\% drop); $^*$high model variance (CV $>$ 50\%), classified by median; s-group's 83.6\% validation coverage is below the 90\% target---sub-nominal non-randomized APS coverage at $K=459$ is expected~\citep{ding2023class}; $^\dagger$meets At-risk criterion (drop $>$15~pp) under the decision framework (Table~\ref{tab:framework_validation}).
\end{table*}

\subsection{Diagnostic Analysis}

Two factors account for much of the variance:

\textbf{Factor 1: Feature temporal stability.} Tasks using transaction IDs (Jaccard $\approx 0.02$) fail catastrophically; tasks using stable entities (products, organizations; Jaccard $> 0.5$) maintain coverage (Table~\ref{tab:overlap}).

\textbf{Factor 2: Feature importance concentration.} Among tasks with universally low Jaccard, concentration determines vulnerability (Section~\ref{sec:shap_results}).

\begin{table}[t]
\centering
\caption{Feature Overlap for Primary Features.}\label{tab:overlap}
\small
\begin{tabular}{lccc}
\toprule
Task & Primary Feature & Jaccard & Drop \\
\midrule
s-shipcond & SALESDOC (ID) & 0.02 & 71.6\% \\
s-group & SALESDOC (ID) & 0.02 & 71.2\% \\
i-incoterms & PRODUCT, PARTY & 0.58 & 11.3\% \\
s-office & SALESORG & 0.61 & 0.1\% \\
\bottomrule
\end{tabular}

\raggedright
\footnotesize
Drop = mean val $-$ mean test coverage (50 seeds).
\end{table}

\subsection{SHAP Concentration Correlates with Failure}\label{sec:shap_results}

\begin{figure}[t]
\centering
\includegraphics[width=\linewidth]{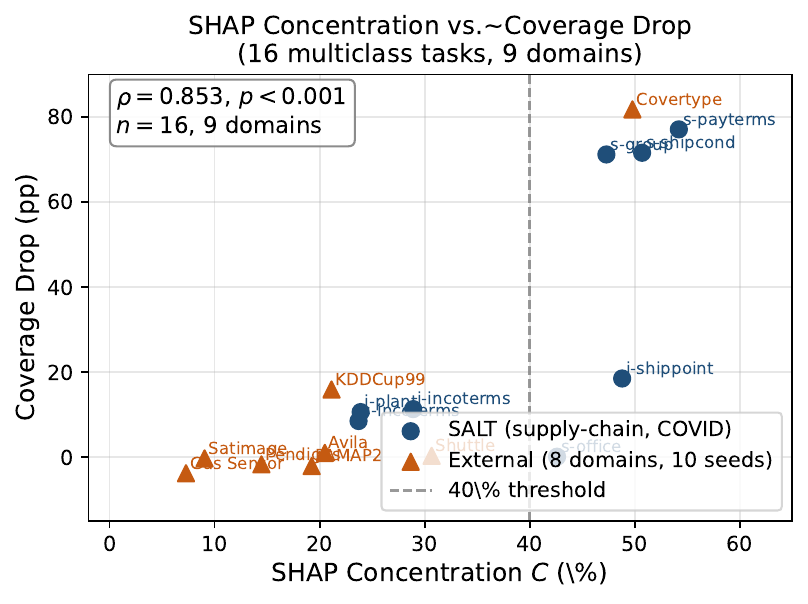}
\caption{\textbf{SHAP Concentration Correlates with Coverage Degradation (Primary Result).} Spearman $\rho=0.853$, $p<0.001$ across all 16 multiclass tasks in 9 domains (dark circles: 8 SALT supply-chain tasks; orange triangles: 8 external domains, 10-seed means). The 40\% threshold (dashed) is derived from validation data only. Within SALT alone: $\rho=0.833$, $p=0.010$ ($n=8$).}\label{fig:n16_correlation}
\end{figure}

Across all 8 SALT multiclass tasks, SHAP concentration correlates with coverage degradation: Spearman $\rho = 0.833$, $p = 0.010$ (Table~\ref{tab:stratified_correlation}, Figure~\ref{fig:n16_correlation}). Bootstrap 95\% CI: $[0.30, 1.00]$ (10,000 resamples, percentile method; BCa would yield a modestly tighter lower bound at $n=8$ but does not change the qualitative conclusion). The correlation magnitude is stable under leave-one-out analysis ($\rho \in [0.75, 0.96]$), though statistical significance is not maintained for 2 of 8 jackknife samples ($p = 0.052$) due to reduced power at $n=7$. SHAP concentration values are highly stable: bootstrap resampling (1,000 resamples of 10K validation samples) yields coefficient of variation below 1\% for all tasks, with 95\% CIs within $\pm$1 percentage point.

\begin{figure*}[t]
\centering
\includegraphics[width=0.85\textwidth]{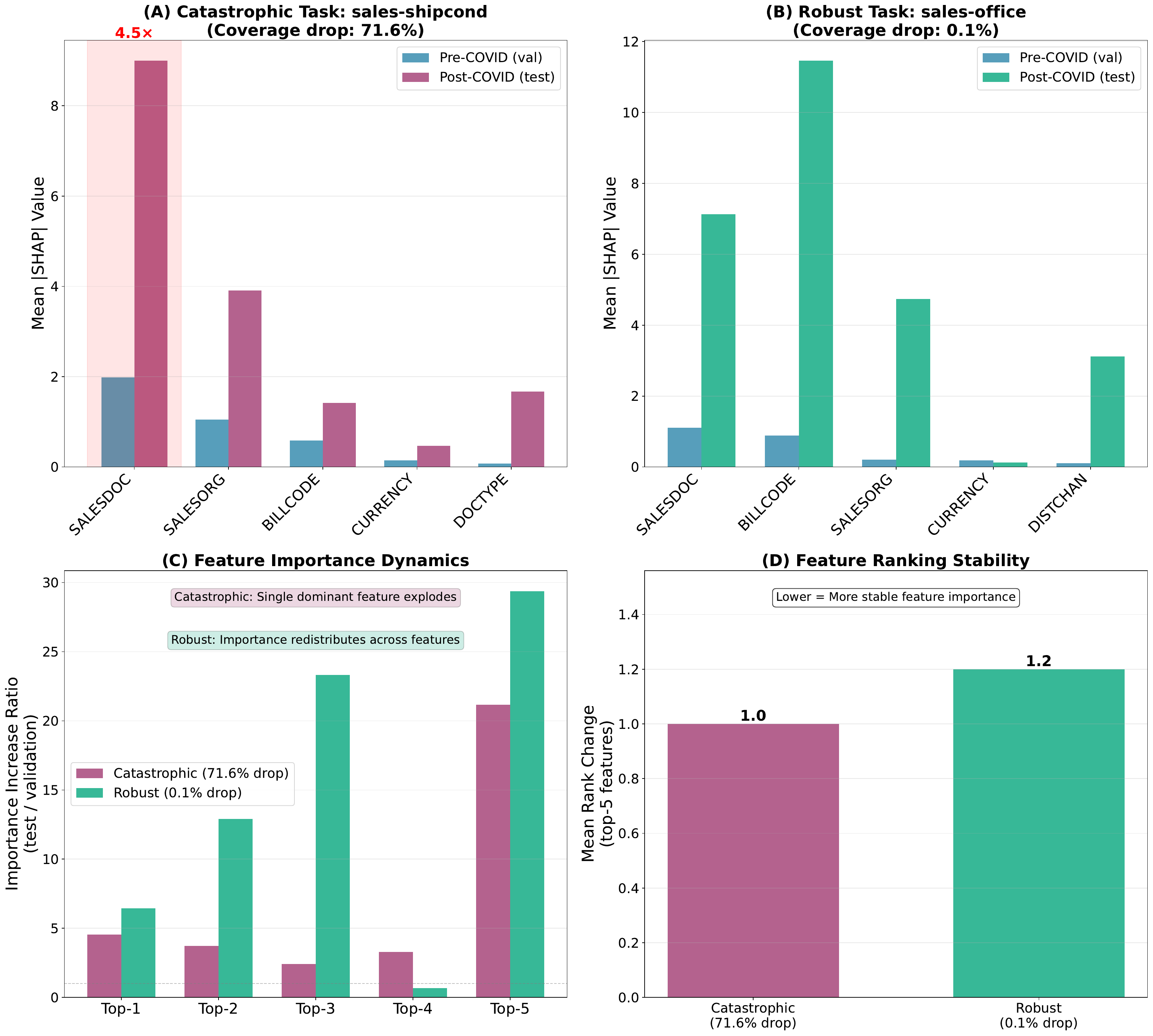}
\caption{\textbf{Feature Importance Dynamics Under Shift.} (A) Catastrophic task: dominant feature explodes 4.5$\times$ while maintaining top rank (concentrated dependence). (B) Robust task: importance redistributes across features with ranking reshuffle (diversified dependence). (C--D) Importance dynamics and ranking stability. Despite $\sim$0\% Jaccard similarity, coverage drops range from negligible to catastrophic.}\label{fig:shap}
\end{figure*}

\textbf{Mechanism.} Figure~\ref{fig:shap} reveals the distinction. Catastrophic tasks: the dominant feature experiences a concentrated importance explosion (4.5$\times$ increase). Robust tasks: importance redistributes across features with ranking reshuffle (10--20$\times$ increases distributed). This matches the prediction of Theorem~\ref{thm:score_inflation}.

\textbf{False positives.} Two tasks with concentration $>$40\% remain robust: sales-office (secondary feature SALESORG has Jaccard $= 0.61$) and item-shippoint (high model variance). These motivate the protective-factor check in our framework (Section~\ref{sec:framework}). The Jaccard $>$ 0.5 and importance $>$ 15\% protective-factor rule is derived from $n=1$ observation (sales-office) and should be treated as provisional until validated on additional false-positive cases.

\textbf{Task independence.} Intraclass correlation analysis confirms task identity explains the majority of seed-level variance in coverage drops (ICC $= 0.675$, 95\% CI $[0.47, 0.90]$; effective $n_{\text{eff}} = 11.7$). Per-task ICC(val,test) is negative for all 8 tasks, indicating paired tests are conservative. Details in Appendix~\ref{app:icc}.

\textbf{Statistical power.} At $n=8$, power to detect $\rho = 0.833$ is $\approx 0.76$ (below the 0.80 convention); the within-SALT result is therefore \textit{exploratory}. The $n=16$ cross-domain result ($\rho=0.853$, $p<0.001$; power $>0.99$) is the confirmatory endpoint.

\textbf{Cross-domain extension.} Extending to 16 multiclass tasks across 9 domains---including 8 external datasets spanning documented geographic, temporal, and benchmark-partition shifts (Section~\ref{sec:cross_domain}; shift mechanisms for 5 UCI pre-defined-split datasets are not separately documented)---yields $\rho=0.853$, $p < 0.001$ (Kendall $\tau = 0.667$; bootstrap 95\% CI $[0.50, 0.96]$). We note that the 40\% threshold was derived from the SALT subset ($n=8$); the $n=16$ correlation therefore includes the in-sample SALT tasks ($\rho=0.833$ within SALT alone) alongside the held-out external 8 datasets on which the threshold was not tuned. The external datasets provide the cleanest out-of-sample validation: Covertype is correctly flagged as catastrophic (81.8~pp drop, 10/10 seeds), 6 deterministic low-concentration datasets remain robust (10/10 seeds each), and KDDCup99 is the principal intermediate-regime false negative.

\begin{table}[t]
\centering
\caption{Stratified Correlation Analysis.}\label{tab:stratified_correlation}
\resizebox{\columnwidth}{!}{%
\begin{tabular}{@{}lccccc@{}}
\toprule
Group & $n$ & $\rho$ & Kendall $\tau$ & $p$($\rho$) & Boot.\ 95\% CI \\
\midrule
Multiclass (SALT)    & 8 & 0.833 & 0.714 & 0.010 & $[0.30, 1.00]$ \\
Multiclass (4 dom.) & 11 & 0.909$^\dagger$ & 0.782 & $<$0.001 & $[0.61, 1.00]$ \\
Multiclass (8 dom.) & 15 & 0.882$^\ddagger$ & 0.714 & $<$0.001 & $[0.60, 0.97]$ \\
\textbf{Multiclass (9 dom.)} & \textbf{16} & \textbf{0.853} & \textbf{0.667} & $<$\textbf{0.001} & $\mathbf{[0.50, 0.96]}$ \\
Combined (10 dom.)$^\S$ & 17 & 0.654 & --- & 0.004 & --- \\
\bottomrule
\end{tabular}}

\raggedright
\footnotesize
LOO stability (SALT): $\rho \in [0.75, 0.96]$; 6/8 jackknife significant. Multiclass (9 domains) is the primary result, excluding binary ceiling-effect tasks. Under the at-risk criterion (drop $>$15~pp), threshold performance at 40\% is Precision $= 0.83$, Recall $= 0.83$ (Appendix~\ref{app:threshold}).
$^\dagger$Single-seed external values; multi-seed-consistent value at $n=11$ is $\rho=0.818$, $p=0.002$ (see Appendix~\ref{app:icc}).
$^\ddagger$$n=15$ is an intermediate analysis (7 external multiclass domains); $n=16$ adds KDDCup99 as the 8th external domain to complete the 9-domain primary endpoint. Stack~Overflow (3 classes, near-binary ceiling effect) is excluded from all multiclass endpoints.
$^\S$Combined row includes all 17 tasks (8 SALT + 9 external, 10 domains: SALT $+$ 9 external), adding Stack~Overflow back to the $n=16$ primary endpoint. Spearman $\rho=0.654$, $p=0.004$ (from Appendix~\ref{app:icc}); Kendall $\tau$ not reported for this auxiliary endpoint.
\end{table}

\subsection{Diagnostic Comparison}\label{sec:baselines}

We compare SHAP concentration against two alternative pre-deployment diagnostics that require no test data (Table~\ref{tab:diagnostic_comparison}).

\textbf{Native feature importance concentration}: top-1 LightGBM gain divided by total gain (zero SHAP cost, single seed; seed=42). Achieves $\rho = 0.667$ ($p = 0.071$)---directionally correct but not significant at $\alpha = 0.05$. The weaker discrimination arises because native gain conflates feature cardinality with predictive importance: sales-office shows 66.1\% native concentration (vs.\ 42.6\% SHAP) despite near-zero coverage drop.

\textbf{Ensemble disagreement}: standard deviation of validation coverage across 50 seeds. Achieves $\rho = 0.452$ ($p = 0.260$). Disagreement captures model instability but not the feature-level mechanism driving conformal failure.

\textbf{Class cardinality as confound.} Within SALT, $\log(\text{num\_classes})$ alone correlates with coverage drop ($\rho = 0.743$, $p = 0.035$), and both partial correlations are non-significant at $n=8$ ($\rho_{\text{partial}} = 0.629$, $p = 0.131$ for concentration; $0.334$, $p = 0.464$ for cardinality; Table~\ref{tab:diagnostic_comparison}). This non-significance is expected at $n=8$---within-SALT the two predictors are correlated and power is insufficient for causal separation. Cross-domain evidence provides complementary support: Covertype (7 classes, $C=49.8\%$) fails catastrophically, while higher-cardinality low-concentration datasets such as Avila (12 classes, $C=20.5\%$), PAMAP2 (12, $19.2\%$), and Pendigits (10, $14.5\%$) remain robust---showing that high class count alone does not predict failure. KDDCup99 (11, $21.1\%$) is a seed-dependent intermediate case rather than a clean robust counterexample. At $n=16$ (8 SALT + 8 external multiclass tasks), the partial Spearman controlling for $\log K$ yields $\rho_{\text{partial}}(\text{conc.}) = 0.771$ ($p = 0.0008$) while $\rho_{\text{partial}}(\log K) = -0.010$ ($p = 0.97$), providing strong cross-domain evidence that concentration drives the severity association independently of class count.

\begin{table}[t]
\centering
\caption{Pre-Deployment Diagnostic Comparison ($n=8$). Spearman $\rho$ with 50-seed mean coverage drop. $\rho_{\text{partial}}$: partial Spearman controlling for the other variable in the SHAP--class-count pair.}\label{tab:diagnostic_comparison}
\small
\begin{tabular}{@{}lcccc@{}}
\toprule
Diagnostic & $\rho$ & $p$ & $\rho_{\text{partial}}$ & $p_{\text{partial}}$ \\
\midrule
SHAP concentration & \textbf{0.833} & \textbf{0.010} & 0.629 & 0.131 \\
$\log(\text{num\_classes})$ & 0.743 & 0.035 & 0.334 & 0.464 \\
Native FI concentration & 0.667 & 0.071 & --- & --- \\
Ensemble disagreement & 0.452 & 0.260 & --- & --- \\
\bottomrule
\end{tabular}
\end{table}

We additionally compare against two \textit{post-hoc} baselines computed from 10-seed ACI experiments:

\textbf{Prediction entropy change}: $\Delta H = H_{\text{test}} - H_{\text{val}}$, where $H = -\sum_c p_c \log p_c$ averaged over instances.

\textbf{Expected Calibration Error change}: \mbox{$\Delta \text{ECE} = \text{ECE}_{\text{test}} - \text{ECE}_{\text{val}}$}.

Both metrics can detect shift \textit{after} test data is observed, but neither is available at deployment time. SHAP concentration is computed solely on validation data, making it actionable for pre-deployment risk assessment. Comparison across 7 tasks in Appendix~\ref{app:baselines}.

\section{Extended Experiments}\label{sec:extended}

\subsection{Adaptive Conformal Inference Across All Tasks}

ACI~\citep{gibbs2021adaptive} recovers marginal coverage on severe tasks (+47.5~pp on s-shipcond, +71.8~pp on s-group, +60.0~pp on s-payterms) but inflates prediction sets; only s-shipcond remains deployable post-adaptation (usable-set rate 30\%: sets $\approx$30\% of $K$=45 classes; s-group inflates to 74--89\% of 459). Robust tasks show minimal benefit. ACI can recover coverage but often sacrifices informativeness in the highest-risk regimes.

\subsection{Alternative Conformal Scoring}\label{sec:raps}

Regularized Adaptive Prediction Sets (RAPS)~\citep{angelopoulos2021uncertainty} shows mechanism-dependent benefit: substantially reduces drops for high-cardinality tasks (s-group: $73.5\%\rightarrow10.4\%$; s-payterms: $79.9\%\rightarrow35.1\%$) but not the most concentrated task (s-shipcond: $60.4\%\rightarrow67.8\%$). This matches Section~\ref{sec:theory}: regularization controls class-accumulation effects, not concentrated single-feature dependence (Appendix~\ref{app:raps}; RAPS used random 50/50 calibration splits vs.\ deterministic main-experiment splits---within-experiment comparisons are unaffected).

\subsection{Comparison with Standard Shift Detection}\label{sec:shift_detection}

MMD, C2ST, and PSI detect shift on all 8 SALT tasks (MMD $p<0.002$, C2ST $>99.9\%$, PSI $>0.002$) but show $\rho \leq 0.19$ with coverage drop---confirming shift detection does not predict failure severity.

\subsection{Retraining Analysis}

Quarterly retraining improves s-shipcond by +18.9~pp ($p=0.036$, unadjusted, single-seed; Holm-corrected $p=0.11$ over 3 severe tasks), partially improves s-payterms (+15.5~pp, non-significant), and is ineffective for s-group ($0.0\%\rightarrow2.2\%$). Improvement concentrates in the final two months of the window; robust tasks remain near-ceiling without intervention.

\subsection{Cross-Domain and External Validation}\label{sec:cross_domain}

External validation on 9 non-supply-chain datasets~\citep{dua2017uci} with the same pipeline (Appendix~\ref{app:external_data}) replicates the pattern: $\rho=0.853$, $p<0.001$ (Kendall $\tau=0.667$) across $n=16$ multiclass tasks. The 8 external multiclass datasets comprise 4 with documented real-world shifts (Covertype geographic, Gas Sensor temporal, KDDCup99 attack-distribution, PAMAP2 subject-based) and 4 null-shift controls (Shuttle, Avila, Pendigits, Satimage: random/pre-defined splits, no genuine shift). The controls are informative: the framework predicts robust coverage for low-concentration models absent genuine shift, confirmed in all 4 cases (10/10 seeds each). Among documented-shift datasets, Covertype ($C=49.8\%$, 81.8~pp drop, 10/10 seeds) is the key catastrophic external case; KDDCup99 ($C=21.1\%$, drop $=15.9\pm21.4$~pp) is the principal intermediate-regime false negative.

\subsection{Placebo Test}

A pre-COVID temporal placebo (train 2018, validate 2019-H1, test 2019-H2) shows 6--143$\times$ lower degradation than the COVID split (Appendix~\ref{app:placebo}), confirming the diagnostic is most informative under severe event-driven shift; the 1--2 order-of-magnitude contrast holds despite the asymmetric training window.

\section{Decision Framework}\label{sec:framework}

This exploratory protocol guides practitioners deploying conformal prediction under potential shift:

\begin{enumerate}
    \item \textbf{Compute} $C = \phi_1 / \sum_j \phi_j$ (Eq.~\ref{eq:concentration}). If $C \leq 40\%$: lower observed risk. If $C > 40\%$: check protective factors---secondary features with train-validation Jaccard $> 0.5$ and importance $> 15\%$ (thresholds empirically derived from a single observed case; treat as provisional); absent these, flag as \textbf{vulnerable}. If $C \in [35\%, 45\%]$: treat as \textbf{uncertain risk} with monitoring regardless of other factors.
    \item \textbf{If vulnerable}: consider quarterly retraining (suggestive effect; Holm-corrected $p=0.11$; single-seed; Appendix~\ref{app:framework}).
    \item \textbf{Monitor} empirical coverage continuously; defer retraining for lower-risk tasks.
\end{enumerate}

\section{Conclusion}\label{sec:conclusion}

Catastrophic tasks become \textit{confidently wrong}---prediction sets shrink (rightward score shift, Appendix~\ref{app:ks}) while coverage collapses, driven by concentrated OOD-feature dependence that generic shift detectors cannot distinguish from robust outcomes. SHAP concentration provides a pre-deployment signal: $\rho=0.853$ ($p<0.001$) across 16 multiclass tasks vs.\ $\rho \leq 0.19$ for MMD/C2ST. Mitigation is mechanism-dependent (RAPS helps class-accumulation failures; retraining/ACI recover coverage). The 40\% threshold is exploratory ($n=8$ SALT tasks) and requires prospective validation beyond boosting models.
Specifically, the diagnostic captures concentrated single-feature dependence (GBM $\rho=0.833$) but not global-sensitivity failures where coverage degrades via coordinated multi-feature sensitivity (MLP $\rho=0.43$, $p=0.29$).

\begin{contributions}
    C.L. conceived the research question, designed and executed all experiments, performed the statistical analysis, and wrote the paper.
\end{contributions}

\begin{acknowledgements}
    We thank the RelBench team for providing the SALT dataset and the anonymous reviewers for their constructive feedback. Code for all experiments and figure generation is submitted as supplementary material.
\end{acknowledgements}

\bibliography{references}

\newpage
\onecolumn

\title{Diagnosing Conformal Prediction Failures Under Distribution Shift:\\A COVID-19 Case Study (Supplementary Material)}
\maketitle

\appendix

\section{Reproducibility Details}\label{app:reproducibility}

All experimental code---including SHAP concentration computation, conformal prediction evaluation, statistical tests, and figure generation---is submitted as supplementary material.

\subsection{LightGBM Hyperparameters}

All models trained with fixed LightGBM~\citep{ke2017lightgbm} settings: objective=multiclass, boosting=gbdt, num\_leaves=31, learning\_rate=0.05, feature\_fraction=0.8, bagging\_fraction=0.8, bagging\_freq=5, num\_boost\_round=500, early\_stopping\_rounds=50, seeds 42--91 (50 seeds). No task-specific tuning was performed.

\subsection{Conformal Prediction}

Method: APS~\citep{romano2020classification}. Target coverage: $(1-\alpha)=0.9$ (90\%). Calibration: deterministic first-half/second-half split of the validation set (i.e., the first $\lfloor n_{\text{val}}/2 \rfloor$ records form the calibration set). This preserves temporal order within the validation period but is not expected to affect the rank-order concentration--drop correlation, as the same split is applied uniformly across all tasks. Quantile: $q = \min(\lceil(n+1)(1-\alpha)\rceil/n, 1)$.

\noindent\textbf{Calibration and test set sizes.} Sales-level tasks (s-shipcond, s-payterms, s-group, s-incoterms, s-office): $n_{\text{val}} = 71{,}474$, $n_{\text{cal}} = 35{,}737$, $n_{\text{test}} = 88{,}942$. Item-level tasks (i-plant, i-shippoint, i-incoterms): $n_{\text{val}} = 293{,}896$, $n_{\text{cal}} = 146{,}948$, $n_{\text{test}} = 402{,}855$. At these calibration set sizes, the finite-sample correction is negligible: $q - (1-\alpha) < 3 \times 10^{-5}$ for all tasks, so finite-sample effects do not contribute to the observed coverage gaps.

\subsection{SHAP Computation}

Method: TreeExplainer~\citep{lundberg2020local}. Subsample: 10,000 per split. Aggregation: mean absolute SHAP across classes. Concentration metric (Eq.~\ref{eq:concentration}): computed on validation data only.

\subsection{Baseline Diagnostics}

Native feature importance concentration uses a single model (seed=42) with LightGBM's gain-based importance; concentration is top-1 gain divided by total gain. Ensemble disagreement is the standard deviation of validation coverage across 50 seeds. Both are computed without SHAP.

\subsection{Computational Resources}

Standard CPU (8 cores, 8GB RAM). Wall-clock: $\sim$3--4 hours for full 8-task, 50-seed suite. Software: Python 3.9, LightGBM 3.3, SHAP 0.41.

\subsection{External Dataset Protocols}\label{app:external_data}

Table~\ref{tab:external_datasets} documents the train/test split, shift mechanism, and dataset characteristics for all 9 external validation datasets. All use the same LightGBM+APS pipeline as SALT (Appendix~\ref{app:reproducibility}), with 10 seeds and no hyperparameter tuning.

\begin{table}[htbp]
\centering
\caption{External Dataset Protocols. Cal = calibration set (50\% of train); $n_\text{tr}$, $n_\text{te}$ = approximate train/test sizes used; Seeds = 10 for all external datasets. Type: \textbf{DS} = documented real-world shift; \textbf{NC} = null-shift control (random/pre-defined split, no genuine shift; low concentration and robust coverage are the expected and observed outcome).}\label{tab:external_datasets}
\small
\begin{tabular}{@{}lp{4.8cm}clllc@{}}
\toprule
Dataset & Shift mechanism & Type & $K$ & $n_\text{tr}$ & $n_\text{te}$ & Source \\
\midrule
Covertype       & Geographic: areas 1--2 (train), area 3 (val/cal), area 4 (test) & DS & 7 & 290,680 & 36,968 & \citep{dua2017uci} \\
Gas Sensor      & Temporal batch drift: batches 1--6 (train), 7--8 (cal), 9--10 (test) & DS & 6 & 9,840 & 4,070 & \citep{dua2017uci} \\
KDDCup99        & Attack-type distribution: official KDD~Cup 1999 train/test files & DS & 11 & 395,073 & 98,769 & \citep{dua2017uci} \\
PAMAP2          & Population shift: subject-based split (unseen subjects at test) & DS & 12 & 1,674,379 & 268,493 & \citep{dua2017uci} \\
Shuttle         & Null-shift control: UCI Statlog; 60/20/20 random split & NC & 7 & 46,400 & 11,600 & \citep{dua2017uci} \\
Avila Bible     & Null-shift control: copyist identification; 50/50 random split & NC & 12 & 10,430 & 10,437 & \citep{dua2017uci} \\
Pendigits       & Null-shift control: UCI pre-defined split (digit recognition) & NC & 10 & 7,494 & 3,498 & \citep{dua2017uci} \\
Satimage        & Null-shift control: satellite image; 60/20/20 random split & NC & 6 & 5,144 & 1,286 & \citep{dua2017uci} \\
Stack Overflow  & Category shift (3-class subset); excluded from $n=16$ primary endpoint (near-binary ceiling effect) & --- & 3 & 65,105 & 127,370 & \citep{dua2017uci} \\
\bottomrule
\end{tabular}

\raggedright
\footnotesize
For Covertype, train uses wilderness areas 1--2 (Spruce-Fir and Lodgepole Pine); conformal calibration uses area 3 (Ponderosa Pine); test uses area 4 (Cottonwood/Willow). For Gas Sensor, each batch contains $\sim$1{,}000 samples across 6 gas types; temporal sensor drift is the documented shift mechanism; $n_\text{tr}$ is batches 1--8 (train + conformal cal); $n_\text{te}$ is batches 9--10. For KDDCup99, classes are the 11 attack categories present in the 10\% subset; the official train file ($\sim$395K) is split 75/25 into LightGBM training and conformal calibration; $n_\text{tr}$ reports total official training file; $n_\text{te}$ is the official test file. For PAMAP2, a subject-based protocol yields a large training pool ($\sim$1.67M samples); $n_\text{tr}$ = LightGBM train + conformal cal combined. For Pendigits, the official UCI train partition (7,494) is split 80/20 into LightGBM training and conformal calibration. For Shuttle, Avila, Satimage, and Stack~Overflow, the full dataset is split 60/20/20 into LightGBM train/conformal cal/test; $n_\text{tr}$ = train+cal pool. SHAP concentration is computed on the conformal calibration partition (validation data) only. Null-shift controls (NC) serve as low-concentration baselines: the framework predicts robust coverage absent genuine shift, which is confirmed in all 4 cases (10/10 seeds each).
\end{table}

\section{Placebo Test Details}\label{app:placebo}

\begin{table}[htbp]
\centering
\caption{Placebo Test: Pre-COVID vs COVID Degradation (50-Seed Mean Drops).}\label{tab:placebo}
\begin{tabular}{lccc}
\toprule
Task & Placebo & COVID & Ratio \\
\midrule
s-shipcond & 0.5\% & 71.6\% & 143$\times$ \\
s-group & 2.0\% & 71.2\% & 36$\times$ \\
s-payterms & 0.0\% & 77.1\% & $>$100$\times$ \\
i-plant & 1.8\% & 10.6\% & 6$\times$ \\
i-shippoint & 1.5\% & 18.5\% & 12$\times$ \\
s-incoterms & 1.2\% & 8.5\% & 7$\times$ \\
i-incoterms & 0.6\% & 11.3\% & 19$\times$ \\
s-office & 0.1\% & 0.1\% & 1$\times$ \\
\bottomrule
\end{tabular}
\end{table}

\section{High Model Variance Analysis}\label{app:variance}

Tasks marked with $^*$ in Table~\ref{tab:main_results} exhibit extreme model variance (CV $>$ 50\%):
\begin{itemize}
    \item \textbf{s-group}: Mean 12.4\%, median \textbf{0.5\%}. Most models fail; outliers inflate mean.
    \item \textbf{i-shippoint}: Mean 72.7\%, median \textbf{89.9\%}. Most models succeed; failures deflate mean.
\end{itemize}

\section{Threshold Sensitivity Analysis}\label{app:threshold}

Table~\ref{tab:threshold_sensitivity} shows threshold performance across $n=16$ multiclass tasks. Note that the 40\% threshold is derived from the SALT concentration gap (in-sample); the 8 external datasets are fully out-of-sample. On the external subset alone at 40\%: Precision~$=$~1.00, Recall~$=$~0.50 (1 TP: Covertype; 1 FN: KDDCup99; 6 TN). Low external recall reflects sparse catastrophic cases; the primary claim rests on the $n=16$ correlation ($\rho=0.853$) rather than threshold classification. Ground-truth labels use the at-risk rule (mean coverage drop $>$15~pp, consistent with 50-seed means in Table~\ref{tab:main_results}). For high-variance tasks, the mean and median labels can differ: i-shippoint has mean drop 18.5~pp (At-risk) but median drop 1.2~pp (robust); Table~\ref{tab:main_results} labels it ROB by median for descriptive purposes, while the framework uses the mean-based threshold for threshold evaluation.

\begin{table}[htbp]
\centering
\caption{Threshold Sensitivity for SHAP Concentration Across 16 Multiclass Tasks (8 SALT + 8 external; Step 2 only, no protective-factor check). At-risk defined as coverage drop $>$15\%.}\label{tab:threshold_sensitivity}
\small
\begin{tabular}{lcccccc}
\toprule
Threshold & Prec & Rec & F1 & TP & FP & FN \\
\midrule
25\% & 0.63 & 0.83 & 0.71 & 5 & 3 & 1 \\
30\% & 0.71 & 0.83 & 0.77 & 5 & 2 & 1 \\
35\% & 0.83 & 0.83 & 0.83 & 5 & 1 & 1 \\
\textbf{40\%} & \textbf{0.83} & \textbf{0.83} & \textbf{0.83} & \textbf{5} & \textbf{1} & \textbf{1} \\
45\% & 1.00 & 0.83 & 0.91 & 5 & 0 & 1 \\
50\% & 1.00 & 0.33 & 0.50 & 2 & 0 & 4 \\
\bottomrule
\end{tabular}

\raggedright
\footnotesize
At 40\%, the single FP is s-office (concentration 42.6\%) which has a protective secondary feature (SALESORG Jaccard $= 0.61$), and the single FN is KDDCup99 (intermediate regime; $C=21.1\%$, drop $=15.9$~pp). At 30\%, the two FPs are s-office and Shuttle ($C=30.7\%$); at 25\%, i-incoterms ($C=28.9\%$) is additionally flagged. Applying the protective-factor check at 40\% removes s-office (precision $= 1.00$) but leaves the KDDCup99 FN (recall $= 0.83$).
\end{table}

\subsection{Leave-One-Out Cross-Validation}\label{app:loocv}

To assess whether the 40\% threshold generalizes beyond post-hoc selection, we perform leave-one-out cross-validation on the SALT dataset ($n=8$). For each fold, we exclude one task, compute the optimal threshold from the remaining 7 tasks, and predict the held-out task's vulnerability status.

\textbf{Summary.} LOO-CV achieves 87.5\% accuracy (7/8 correct predictions). The threshold remains stable across folds: $43.1\% \pm 5.0\%$ (mean $\pm$ std). The effect size separating failed from robust tasks is Cohen's $d = 3.08$ (large), with bootstrap 95\% CI $[0.31, 1.00]$ for the underlying correlation.

\begin{table}[htbp]
\centering
\caption{LOO-CV Per-Fold Predictions (SALT, $n=8$). Threshold = optimal threshold from 7 training tasks; Pred = predicted status at that threshold; Actual = observed outcome (Failed if drop $>$15~pp).}\label{tab:loocv}
\small
\begin{tabular}{@{}lcccccc@{}}
\toprule
Task & Conc. (\%) & Drop (pp) & Thresh (\%) & Pred & Actual & Correct \\
\midrule
s-shipcond & 50.7 & 71.6 & 45 & Vuln & Failed & \checkmark \\
s-group & 47.3 & 71.2 & 45 & Vuln & Failed & \checkmark \\
s-payterms & 54.2 & 77.1 & 45 & Vuln & Failed & \checkmark \\
i-plant & 23.9 & 10.6 & 45 & Robust & Robust & \checkmark \\
i-shippoint & 48.8 & 18.5 & 45 & Vuln & Failed & \checkmark \\
s-incoterms & 23.7 & 8.5 & 45 & Robust & Robust & \checkmark \\
i-incoterms & 28.9 & 11.3 & 45 & Robust & Robust & \checkmark \\
\textbf{s-office} & \textbf{42.6} & \textbf{0.1} & \textbf{30} & \textbf{Vuln} & \textbf{Robust} & $\times$ \\
\bottomrule
\end{tabular}
\end{table}

\textbf{Effect size analysis.} Failed tasks ($n=4$) have mean concentration $50.2\% \pm 2.6\%$, while robust tasks ($n=4$) have mean concentration $29.8\% \pm 7.7\%$. The separation of 20.4~pp yields Cohen's $d = 3.08$ (large effect), confirming that concentration reliably distinguishes vulnerability classes.

\textbf{Misclassification analysis.} The single misclassification (s-office) is a meaningful boundary case rather than a diagnostic failure. Sales-office has high concentration (42.6\%) but maintains near-perfect coverage (0.1~pp drop). The protective factor is high Jaccard overlap: the dominant feature (SALESORGANIZATION) has Jaccard $= 0.61$ between train and test, meaning the dominant feature's value distribution is stable across the temporal shift. When the dominant feature does \textit{not} shift out of distribution, high concentration does not imply vulnerability---the model's concentrated dependence remains valid. This motivates the protective-factor check in the decision framework (Section~\ref{sec:framework}): high concentration is a necessary but not sufficient condition for vulnerability; stable dominant features can protect concentrated models from failure.

\section{Decision Framework Validation}\label{app:framework}

\begin{table}[htbp]
\centering
\caption{Concentration Check Applied to All Multiclass Tasks (Actual = At-risk if drop $>$15~pp).}\label{tab:framework_validation}
\small
\begin{tabular}{@{}lcccc@{}}
\toprule
Task & Conc. (\%) & Step 2 & Actual ($>$15~pp) & Note \\
\midrule
\multicolumn{5}{l}{\textit{SALT (in-sample)}} \\
s-payterms & 54.2 & VULN & At-risk & --- \\
s-shipcond & 50.7 & VULN & At-risk & --- \\
i-shippoint & 48.8 & VULN & At-risk$^*$ & High variance$^\dagger$ \\
s-group & 47.3 & VULN & At-risk$^*$ & --- \\
s-office & 42.6 & VULN & ROB & Protective$^\ddagger$ \\
i-incoterms & 28.9 & ROB & ROB & --- \\
i-plant & 23.9 & ROB & ROB & --- \\
s-incoterms & 23.7 & ROB & ROB & --- \\
\midrule
\multicolumn{5}{l}{\textit{External (held-out)}} \\
Covertype & 49.8 & VULN & At-risk & 10/10 seeds \\
Shuttle & 30.7 & ROB & ROB & 9/10 seeds \\
Avila Bible & 20.5 & ROB & ROB & 10/10 seeds \\
PAMAP2 & 19.2 & ROB & ROB & 10/10 seeds \\
KDDCup99 & 21.1 & ROB & At-risk$^\S$ & Seed-dependent (FN) \\
Pendigits & 14.5 & ROB & ROB & 10/10 seeds \\
Satimage & 9.0 & ROB & ROB & 10/10 seeds \\
Gas Sensor & 7.3 & ROB & ROB & 10/10 seeds \\
Stack Overflow & 48.9 & VULN & ROB & Near-binary ceiling$^\P$ \\
\bottomrule
\end{tabular}

\raggedright
\footnotesize
$^\dagger$Median drop 1.2\%; most seeds maintain coverage.
$^\ddagger$SALESORGANIZATION: train-validation Jaccard$=$0.61, importance$>$15\%.
$^\S$Concentration below threshold ($C=21.1\pm7.5\%$), but mean drop exceeds the at-risk cutoff (15.9~pp); outcome seed-dependent (range $-$0.8 to 73.5~pp).
In-sample F1 $= 0.89$ (1 FP: s-office). External: Covertype deterministic catastrophic (10/10); five low-concentration datasets deterministic robust (10/10); Shuttle near-deterministic (9/10); KDDCup99 is the intermediate-regime false negative.
$^\P$Stack Overflow ($K=3$, $C=48.9\%$): high concentration predicts VULN, but near-binary class structure creates a ceiling effect (only 3 classes; APS prediction sets trivially remain small). Actual coverage is robust ($-7.0$~pp). Excluded from $n=16$ multiclass primary endpoint; confirms diagnostic applies to multiclass ($K>3$) settings only.
\end{table}

\noindent\textbf{Seed stability protocol.} KDDCup99's multi-seed analysis ($C=21.1\pm7.5\%$) reveals that single-seed SHAP concentration can be unstable. We recommend computing $C$ over $k \geq 5$ seeds and using the mean. Tasks in the ambiguous zone (35--45\%) should receive additional scrutiny: if the upper 95\% CI exceeds 40\%, treat as potentially vulnerable. This safety override is designed to reduce KDDCup99-like false negatives in intermediate regimes. For SALT tasks, concentration is stable across 50 seeds (within-task CV $<$ 5\%); instability arises primarily when the top-feature identity varies across seeds.

\noindent\textbf{Threshold selection.} The 40\% threshold is proposed based on the empirical distribution of SHAP concentration values on validation data: a natural gap separates low-concentration tasks (24--29\%) from high-concentration tasks (43--54\%), visible in Table~\ref{tab:framework_validation}. We emphasize that this threshold is \textit{exploratory}, derived from $n=8$ multiclass tasks, and should be validated before deployment in new domains.

\noindent\textbf{Cross-domain transfer test.} The 40\% threshold applied without tuning to 9 external domains yields 7/9 deterministic or near-deterministic outcomes (6 at 10/10, Shuttle at 9/10), with Covertype correctly flagged as catastrophic and Gas Sensor correctly unflagged as robust. KDDCup99 is the principal intermediate-regime false negative, while Stack Overflow (3 classes) exhibits the near-binary ceiling effect. Across the $n=16$ multiclass set: $\rho = 0.853$, $p < 0.001$ (Section~\ref{sec:cross_domain}). The sensitivity analysis in Appendix~\ref{app:threshold} reports threshold performance across 30--50\%; given sparse catastrophic external cases, this threshold should be treated as provisional.

\section{Pre-Deployment vs Post-Hoc Baselines}\label{app:baselines}

SHAP concentration is computed on validation data \textit{before} deployment. Prediction entropy and ECE require observing test data. This section reports entropy and ECE changes for completeness, noting that they are not available as pre-deployment diagnostics.

\begin{table}[htbp]
\centering
\caption{Pre-Deployment (SHAP) vs Post-Hoc (Entropy, ECE) Shift Diagnostics (10-Seed ACI Experiments). SHAP concentration is computed on validation data only. $\Delta$Entropy and $\Delta$ECE require test-time observations.}\label{tab:baselines}
\small
\begin{tabular}{@{}lcccc@{}}
\toprule
Task & Cat & SHAP Conc.$^\dagger$ & $\Delta$Entropy & $\Delta$ECE \\
\midrule
\multicolumn{5}{l}{\textit{Severe tasks}} \\
s-shipcond & SEV & 50.7\% & $-$0.746 & +0.408 \\
s-group & SEV & 47.3\% & $-$1.274 & +0.176 \\
s-payterms & SEV & 54.2\% & $-$1.979 & +0.629 \\
\midrule
\multicolumn{5}{l}{\textit{Robust tasks}} \\
i-plant & ROB & 23.9\% & +0.181 & +0.131 \\
i-shippoint & ROB & 48.8\% & +0.469 & +0.073 \\
s-incoterms & ROB & 23.7\% & +0.364 & +0.350 \\
s-office & ROB & 42.6\% & +0.032 & +0.104 \\
\bottomrule
\end{tabular}

\raggedright
\footnotesize
$^\dagger$Pre-deployment diagnostic (computed on validation data only). item-incoterms omitted (3-seed pilot only).
\end{table}

Counter-intuitively, all three catastrophic tasks (coverage drop $>$ 70\%) show \textit{decreasing} prediction entropy ($\Delta H < 0$): the model becomes \textit{more} confident at test time, not less. This occurs because concentrated dependence on an OOD feature produces arbitrary but confident predictions. In contrast, the two robust tasks with non-trivial drops (i-plant, i-shippoint; coverage drops of 10--19\%) show \textit{increasing} entropy ($\Delta H > 0$), consistent with standard distributional broadening. A practitioner monitoring only entropy would correctly detect shift in moderate cases but would be \textit{misled} in the most damaging catastrophic cases---observing decreasing uncertainty while coverage collapses. ECE increases across all tasks but does not discriminate between severity levels (both catastrophic and robust tasks show increases of +0.07 to +0.63). In this SALT comparison, SHAP concentration is the only evaluated pre-deployment metric that flags all three catastrophic tasks.

\section{Task Independence Analysis}\label{app:icc}

We analyze pseudo-replication in the 50-seed design using intraclass correlation (ICC).

\textbf{Between-task ICC.} For coverage drops, ICC $= 0.675$ (95\% CI: $[0.47, 0.90]$; $F(7,392)=104.6$), confirming that task identity explains the majority of seed-level variance. The design effect is 34.1, yielding effective $n_{\text{eff}} = 400/34.1 = 11.7$---exceeding the 8 task-level observations. For test coverages, ICC $= 0.720$ (95\% CI: $[0.52, 0.92]$). These values confirm that the $n=8$ tasks constitute meaningfully independent observations for the Spearman correlation analysis.

\textbf{Per-task ICC(val,test).} For each task, we compute ICC between the 50 validation coverages and 50 test coverages (paired by seed). All 8 per-task ICCs are negative ($-0.80$ to $-0.10$), indicating that seeds producing above-average validation coverage do \textit{not} produce above-average test coverage. This means paired Wilcoxon tests are conservative: the effective sample size for within-task paired tests \textit{exceeds} 50, so the reported $p$-values are, if anything, too large.

\begin{table}[htbp]
\centering
\caption{Per-Task ICC Between Validation and Test Coverages (50 Seeds).}\label{tab:per_task_icc}
\small
\begin{tabular}{@{}lccc@{}}
\toprule
Task & ICC(val,test) & Design Effect & Eff.\ $n$ \\
\midrule
s-payterms  & $-$0.80 & 0.20 & 250 \\
s-shipcond  & $-$0.77 & 0.23 & 220 \\
s-group     & $-$0.72 & 0.28 & 177 \\
i-plant     & $-$0.42 & 0.58 & 86 \\
i-incoterms & $-$0.37 & 0.63 & 80 \\
s-incoterms & $-$0.35 & 0.65 & 77 \\
s-office    & $-$0.15 & 0.85 & 59 \\
i-shippoint & $-$0.10 & 0.90 & 55 \\
\bottomrule
\end{tabular}
\end{table}

\textbf{Multiple comparisons.} We tested 5 concentration metrics (top-1, top-2, top-3, HHI, entropy) against coverage drop. Under Holm--Bonferroni correction for this 5-test family, the adjusted $p$-value for top-1 concentration is $5 \times 0.010 = 0.050$, at the conventional significance boundary. For the primary correlation subsets ($n=8$, $n=11$, $n=16$), all adjusted $p$-values remain significant (Holm-corrected $p \leq 0.021$). We note that top-1 was the a priori natural choice (Section~\ref{sec:why_shap}), not one of 5 equally plausible alternatives; the correction is therefore conservative.

\textbf{Multi-seed aggregation.} The primary $n=16$ correlation ($\rho = 0.853$) uses 10-seed means for all external datasets and 50-seed means for SALT tasks. The early $n=11$ result ($\rho = 0.909$) used single-seed external values; switching to multi-seed means at $n=11$ yields $\rho = 0.818$, $p = 0.002$, with the reduction driven by KDDCup99 whose multi-seed mean drop (15.9~pp) exceeds its single-seed value ($-$0.8~pp). This intermediate-concentration regime ($C=21.1\pm7.5\%$) exhibits high outcome variance characteristic of the boundary between robust and vulnerable. Extending to $n=16$ with consistent multi-seed means yields $\rho = 0.853$ ($p < 0.001$). Including Stack Overflow (3 classes, near-binary ceiling) to make $n=17$ weakens correlation to $\rho = 0.654$.

\section{Theory Details and Score CDFs}\label{app:theory_proof}

\textbf{Proof details.} Starting from~\eqref{eq:aps_equiv}, define $L := \sum_{y \in \mathcal{B}}\hat p(y)$ for classes below the true-label rank. Since each term in $L$ is $<\hat p(y^*)$ and $|\mathcal{B}| \le K-1$, $L \le (K-1)\hat p(y^*)$, so $s=1-L$ yields~\eqref{eq:score_pointwise} after substituting (A1)--(A2). Taking expectations and applying (A3) gives~\eqref{eq:score_bound}; differentiating gives monotonicity in $C$ when $\varepsilon < \bar h$. Rearranging~\eqref{eq:score_pointwise} gives the threshold form in~\eqref{eq:cov_bound}, and differentiating $T(C)$ shows the coverage upper bound decreases with concentration under the theorem condition.

\textbf{Conservative bound verification.} Using $\varepsilon=0$ and $\bar h=1/K$, Theorem~\ref{thm:score_inflation}(ii) yields conservative lower bounds verified on all five applicable tasks: s-shipcond ($0.518$ vs.\ observed $0.98$), s-payterms ($0.545$ vs.\ $0.92$), s-group ($0.474$ vs.\ $1.00$), i-plant ($0.261$ vs.\ $0.86$), and s-incoterms ($0.296$ vs.\ $0.87$). The large gaps between bounds and observed values reflect conservative assumptions in the $(K-1)$ counting bound and the uniform residual prior $\bar{h}=1/K$; tighter bounds follow from empirical $\bar{h}$ estimates.

\begin{figure}[htbp]
\centering
\includegraphics[width=\linewidth]{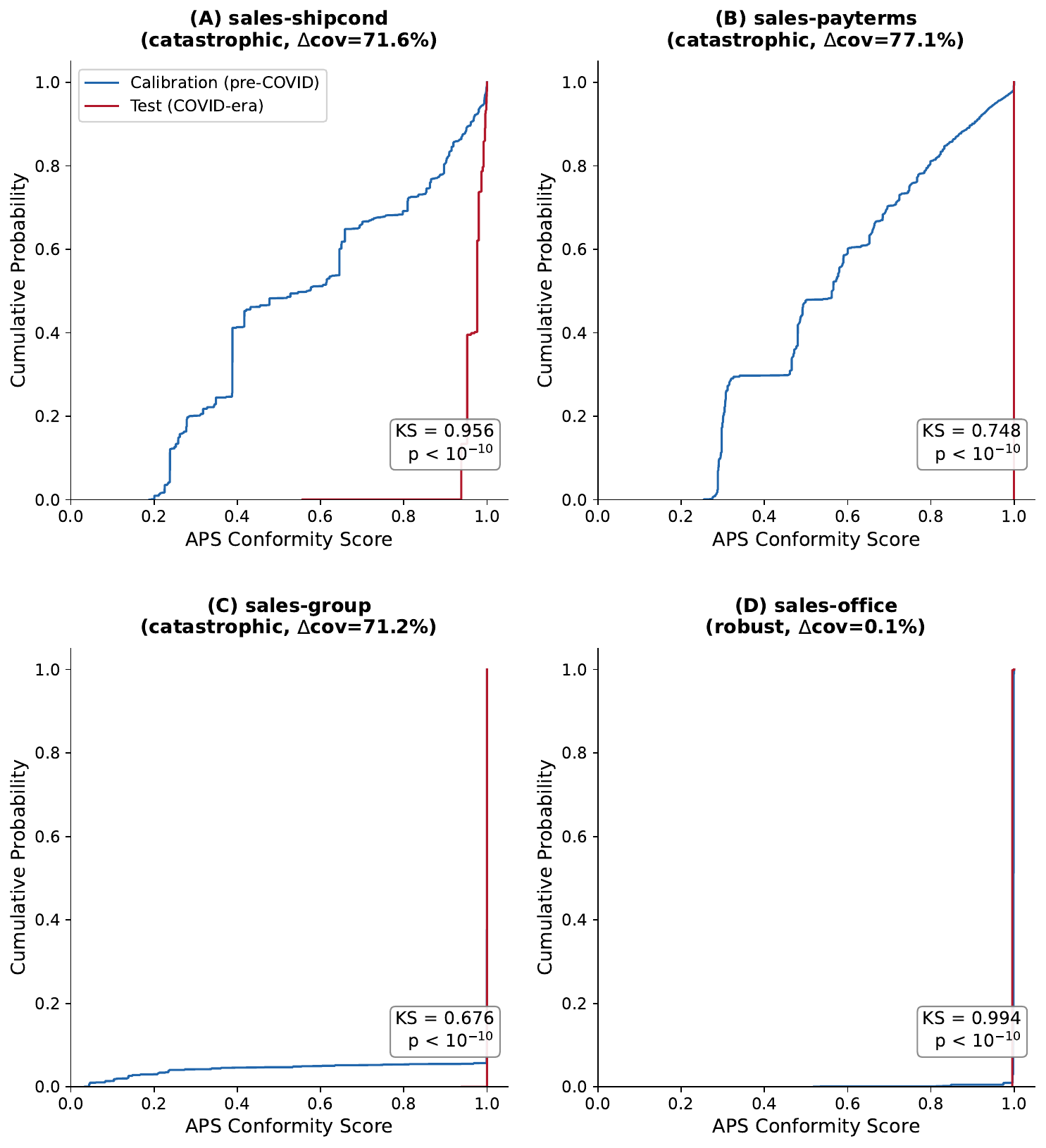}
\caption{\textbf{Conformity score CDFs (seed=42).} Catastrophic tasks show rightward test-score shifts (stochastic dominance), while robust tasks can show near-degenerate or left-shift behavior.}\label{fig:cdfs}
\end{figure}

\section{Conformity Score Stochastic Dominance}\label{app:ks}

Table~\ref{tab:ks} reports two-sample KS statistics comparing calibration and test APS conformity scores (seed=42, non-randomized APS). Higher conformity scores indicate the model requires more cumulative probability mass to reach the true label---i.e., the model is less confident about the correct class.

\begin{table}[htbp]
\centering
\caption{KS Tests: Calibration vs.\ Test Conformity Score Distributions (Seed=42).}\label{tab:ks}
\small
\begin{tabular}{@{}lcccc@{}}
\toprule
Task & KS & $p$ & Mean$_{\text{cal}}$ & Mean$_{\text{test}}$ \\
\midrule
\multicolumn{5}{l}{\textit{Catastrophic tasks (test scores $>$ calibration)}} \\
s-shipcond & 0.956 & $<10^{-10}$ & 0.54 & 0.98 \\
s-payterms & 0.748 & $<10^{-10}$ & 0.65 & 0.92 \\
s-group & 0.676 & $<10^{-10}$ & 0.98 & 1.00 \\
\midrule
\multicolumn{5}{l}{\textit{Robust tasks}} \\
i-plant & 0.741 & $<10^{-10}$ & 0.73 & 0.86 \\
s-incoterms & 0.633 & $<10^{-10}$ & 0.75 & 0.87 \\
s-office & 0.994 & $<10^{-10}$ & 1.00 & 1.00 \\
i-incoterms & 0.559 & $<10^{-10}$ & 0.78 & 0.66 \\
i-shippoint & 0.505 & $<10^{-10}$ & 0.71 & 0.58 \\
\bottomrule
\end{tabular}

\raggedright
\footnotesize
All KS tests are two-sided. Non-randomized APS scores (conservative coverage guarantee). For catastrophic tasks, test scores shift rightward (stochastic dominance), consistent with Theorem~\ref{thm:score_inflation}. Two robust tasks (i-incoterms, i-shippoint) show test scores shifted \textit{leftward}---the model becomes more confident during COVID, possibly due to reduced product diversity.
\end{table}

\section{RAPS Comparison}\label{app:raps}

Table~\ref{tab:raps} compares APS and RAPS coverage drops across all 8 tasks with 10-seed means ($\lambda=0.01$, $k_{\text{reg}}=5$).

\begin{table}[htbp]
\centering
\caption{APS vs.\ RAPS Coverage Drop (10-Seed Means $\pm$ Std). $\lambda=0.01$, $k_{\text{reg}}=5$.}\label{tab:raps}
\small
\begin{tabular}{@{}lccccc@{}}
\toprule
Task & Cl & APS Drop & RAPS Drop & $\Delta$ & Note \\
\midrule
\multicolumn{6}{l}{\textit{Severe tasks}} \\
s-shipcond & 45 & $60.4\pm31.8$ & $67.8\pm25.4$ & +7.4 & No improvement \\
s-group & 459 & $73.5\pm33.9$ & $10.4\pm1.3$ & $-$63.1 & RAPS protects \\
s-payterms & 137 & $79.9\pm29.4$ & $35.1\pm28.6$ & $-$44.8 & Partial recovery \\
\midrule
\multicolumn{6}{l}{\textit{Non-severe tasks}} \\
i-plant & 35 & $11.8\pm9.0$ & $13.1\pm5.8$ & +1.3 & Near-identical \\
i-shippoint & 69 & $9.3\pm27.6$ & $20.5\pm21.6$ & +11.2 & RAPS worse \\
i-incoterms & 13 & $10.8\pm9.1$ & $10.8\pm9.1$ & 0.0 & Identical \\
s-incoterms & 13 & $6.7\pm6.8$ & $6.7\pm6.8$ & 0.0 & Identical \\
s-office & 25 & $0.04\pm0.03$ & $0.04\pm0.03$ & 0.0 & Identical \\
\bottomrule
\end{tabular}

\raggedright
\footnotesize
All values are 10-seed means $\pm$ std. APS drops differ from 50-seed Table~\ref{tab:main_results} due to smaller seed range (10 vs.\ 50) and high variance. The key comparison is APS vs.\ RAPS within the same seed set. Note: RAPS experiments used a random 50/50 calibration split rather than the deterministic first-half/second-half split used in the main APS experiments (Appendix~\ref{app:reproducibility}); this protocol difference does not affect the within-experiment APS vs.\ RAPS comparison but means RAPS coverage values are not directly comparable to Table~\ref{tab:main_results}. RAPS eliminates the catastrophic drop for sales-group by truncating the class accumulation that amplifies score inflation in many-class settings (mean set size: 232/459 classes). Sales-shipcond is unhelped because single-feature dependence places dominant mass on the wrong class \textit{before} the $k_{\text{reg}}=5$ penalty activates---the exchangeability violation occurs at the top of the APS ordering, not through class accumulation. For low-class tasks ($\leq$35 classes), RAPS and APS produce identical or near-identical drops, confirming RAPS regularization has negligible effect when prediction sets are already small. Item-shippoint ($K=69$) is an exception: RAPS worsens coverage by 11.2~pp (paired Wilcoxon $p=0.064$, 10 seeds), reflecting that the regularization penalty can disrupt calibration for intermediate-cardinality tasks with high seed variance.
\end{table}

\section{Model Class Sensitivity}\label{app:rf}

To test whether SHAP concentration is diagnostic across model classes, we replicate the analysis with \texttt{RandomForestClassifier} (500 trees), \texttt{XGBClassifier}, and \texttt{CatBoostClassifier} (ordered boosting), all with seed=42 and the same data pipeline (Table~\ref{tab:rf}).

\begin{table}[htbp]
\centering
\caption{SHAP Concentration and Coverage Drop Across Model Classes.}\label{tab:rf}
\small
\begin{tabular}{@{}lcccccccc@{}}
\toprule
Task & LGB & RF & XGB & CB & LGB & RF & XGB & CB \\
 & \multicolumn{4}{c}{Concentration (\%)} & \multicolumn{4}{c}{Coverage Drop (\%)} \\
\midrule
s-shipcond & 50.7 & 57.4 & 44.6 & 53.7 & 71.6 & 15.8 & 47.2 & 1.1 \\
s-group & 47.3 & 22.5 & 33.3 & 31.8 & 71.2 & 50.0 & 56.8 & 14.5 \\
s-payterms & 54.2 & 27.9 & 30.6 & 25.7 & 77.1 & 13.8 & 20.4 & $-$5.3 \\
i-shippoint & 48.8 & 52.1 & 41.0 & 43.1 & 18.5 & 22.7 & 80.4 & 61.8 \\
i-plant & 23.9 & 26.8 & 26.0 & 29.0 & 10.6 & 15.3 & 9.7 & 8.5 \\
s-incoterms & 23.7 & 26.8 & 21.6 & 28.2 & 8.5 & 2.5 & 3.7 & $-$1.5 \\
i-incoterms & 28.9 & 17.5 & 18.4 & 21.7 & 11.3 & 2.9 & 1.1 & $-$0.4 \\
s-office & 42.6 & 24.5 & 36.8 & 33.5 & 0.1 & 0.0 & 0.0 & $-$0.0 \\
\bottomrule
\end{tabular}
\end{table}

\textbf{Results.} The diagnostic's predictive power varies by model class (Table~\ref{tab:model_sensitivity}). All three gradient-boosted implementations show positive correlation: LightGBM ($\rho = 0.833$, $p = 0.010$), CatBoost ($\rho = 0.667$, $p = 0.071$), and XGBoost ($\rho = 0.548$, $p = 0.16$). RandomForest ($\rho = 0.30$, $p = 0.47$) does not replicate.

\textbf{Mixed-effects analysis.} Since (task, model) pairs share data and temporal split, we fit a linear mixed-effects model with task as a random intercept to properly account for clustering: $\text{Drop}_{ij} = \beta_0 + \beta_1 \cdot \text{Concentration}_{ij} + u_i + \varepsilon_{ij}$, where $u_i \sim \mathcal{N}(0, \sigma^2_u)$ is the task random effect. Across the three gradient-boosted models ($n=24$, 8 tasks $\times$ 3 models), the concentration effect is $\beta_1 = 1.64$ (95\% CI $[0.70, 2.58]$, Wald $p = 0.0006$): each 1\% increase in concentration predicts 1.6~pp additional coverage drop, after accounting for task-level clustering. With only 8 task-level clusters, Wald inference is anti-conservative (random-intercept variance is downward-biased at $<20$ clusters); a Kenward--Roger df correction typically yields $p$ in the range $0.01$--$0.03$ at fewer than 10 clusters (estimated; not computed). The effect size $\beta_1 = 1.64$ (95\% CI $[0.70, 2.58]$) is the primary evidence---the CI excludes zero regardless of the p-value convention. Including model class as a fixed effect yields $\beta_1 = 1.66$. With all four models including RF ($n=32$): $\beta_1 = 1.06$ ($p = 0.010$).

\begin{table}[htbp]
\centering
\caption{Concentration--Drop Relationship by Model Class.}\label{tab:model_sensitivity}
\small
\begin{tabular}{@{}lccc@{}}
\toprule
Model & Method & Conc.--Drop $\rho$ & $p$ \\
\midrule
LightGBM & Boosting (GOSS) & \textbf{0.833} & \textbf{0.010} \\
CatBoost & Boosting (ordered) & 0.667 & 0.071 \\
XGBoost & Boosting (histogram) & 0.548 & 0.16 \\
RandomForest & Bagging & 0.30 & 0.47 \\
\midrule
\multicolumn{2}{@{}l}{Mixed-effects, 3 boosting ($n=24$)} & $\beta_1=1.64$ & \textbf{$<$0.001} \\
\multicolumn{2}{@{}l}{Mixed-effects, all 4 ($n=32$)} & $\beta_1=1.06$ & \textbf{0.010} \\
\bottomrule
\end{tabular}
\end{table}

\textbf{Interpretation.} All three gradient-boosted implementations show the concentration--drop relationship, but with different vulnerability profiles. CatBoost's ordered boosting produces dramatically different task-level vulnerabilities: LGB-catastrophic tasks (s-shipcond: 71.6\% LGB drop) show only 1.1\% CatBoost drop, while i-shippoint reverses (18.5\% LGB $\rightarrow$ 61.8\% CatBoost). This confirms each model's SHAP concentration captures its \textit{own} learned dependence---not a universal data property---and that cross-model transfer is unreliable (LGB conc.\ $\rightarrow$ CatBoost drop: $\rho = 0.07$). RF's bagging produces smoother probability surfaces, compressing both concentration range and vulnerability profiles. Specifically, RF's failure to replicate ($\rho = 0.30$) has two mechanistic causes: (1)~\textit{compressed concentration range}---RF concentrations span 17.5--57.4\% vs.\ LGB's 23.7--54.2\%, but the RF range is driven by a single outlier (s-shipcond at 57.4\%); excluding it, the RF range is 17.5--26.8\%, offering almost no discriminative signal; (2)~\textit{smoothed probability surfaces}---bagging averages 500 independent trees, each seeing random feature subsets, which distributes effective dependence more evenly than boosting's sequential error-correction. The result is that RF rarely develops the concentrated single-feature dependence that drives catastrophic conformal failure in boosting models, and consequently SHAP concentration has limited predictive power for RF vulnerability.

The diagnostic is \textit{model-specific}: it measures how a particular model's learned dependence structure creates vulnerability. Practitioners should compute concentration for their \textit{deployed} model class rather than transferring results across model families.

\textbf{Neural network extension.} To test beyond tree-based models, we evaluate a 2-layer MLP (128/64 hidden units, ReLU, softmax, permutation importance) on all 8 SALT tasks (10 seeds for sales-level tasks, 3 for item-level). The within-MLP concentration--drop correlation is $\rho = 0.43$ ($p = 0.29$), similar to RF and not significant. However, the model-specificity claim is strikingly confirmed across all 8 tasks: the MLP learns fundamentally different concentration profiles from tree-based models. Key examples: (1)~i-shippoint shows MLP $C=77.2\%$, drop $= 82.3$~pp vs.\ LGB $C=48.8\%$, drop $= 18.5$~pp---the MLP concentrates nearly all importance on a single feature and fails catastrophically, while LGB distributes importance and maintains coverage; (2)~s-incoterms shows MLP $C=30.3\%$, drop $= 42.3$~pp vs.\ LGB $C=23.7\%$, drop $= 8.5$~pp---the MLP's greater concentration predicts its greater failure; (3)~s-office shows MLP $C \approx 0\%$, drop $= 0.1$~pp---perfect robustness prediction. The non-significant within-MLP $\rho$ is driven by two tasks (s-group: $C=27.5\%$, drop $= 78.4$~pp; s-payterms: $C=28.0\%$, drop $= 60.6$~pp) where the MLP distributes importance evenly yet still fails catastrophically---suggesting the MLP's failure mechanism involves global sensitivity to the shift rather than concentrated single-feature dependence. This confirms that SHAP concentration is diagnostic specifically for the concentrated-dependence failure mode that characterizes gradient-boosted models, not a universal predictor of all failure modes.

\subsection{MLP vs LightGBM Task-Level Comparison}\label{app:mlp_lgb_comparison}

Table~\ref{tab:mlp_lgb_comparison} provides a structured comparison of SHAP concentration and coverage drop between MLP and LightGBM across all 8 SALT tasks.

\begin{table}[htbp]
\centering
\caption{Task-Level Comparison: MLP vs LightGBM. MLP\_C = MLP SHAP concentration; MLP\_Drop = MLP coverage drop (pp); LGB\_C = LightGBM SHAP concentration; LGB\_Drop = LightGBM coverage drop (pp); Mode = failure mode classification.}\label{tab:mlp_lgb_comparison}
\small
\begin{tabular}{@{}lccccc@{}}
\toprule
Task & MLP\_C & MLP\_Drop & LGB\_C & LGB\_Drop & Mode \\
\midrule
s-group & 27.5\% & 78.4pp & 47.3\% & 71.2pp & Global \\
s-payterms & 28.0\% & 60.6pp & 54.2\% & 77.1pp & Global \\
i-shippoint & 77.2\% & 82.3pp & 48.8\% & 18.5pp & Conc \\
s-incoterms & 30.3\% & 42.3pp & 23.7\% & 8.5pp & Mixed \\
s-shipcond & 53.1\% & 67.1pp & 50.7\% & 71.6pp & Conc \\
i-incoterms & 31.0\% & 24.1pp & 28.9\% & 11.3pp & Mixed \\
i-plant & 28.2\% & 26.7pp & 23.9\% & 10.6pp & Mixed \\
s-office & 0.0\% & 0.1pp & 42.6\% & 0.1pp & N/A \\
\bottomrule
\end{tabular}

\raggedright
\footnotesize
Mode: Conc = concentrated-dependence failure (C $>$ 40\% predicts drop); Global = global-sensitivity failure (C $\leq$ 40\% yet catastrophic drop); Mixed = intermediate regime; N/A = no meaningful drop in either model. Critical observation: s-group and s-payterms show MLP C $\approx$ 28\% yet 60--78pp drops, demonstrating the global-sensitivity failure mode where SHAP concentration does not capture the MLP's vulnerability mechanism.
\end{table}

\subsection{Failure Mode Definitions and Quantitative Criteria}\label{app:failure_modes}

We formalize two distinct failure modes that determine when SHAP concentration is diagnostic.

\smallskip
\noindent\fbox{\parbox{0.97\columnwidth}{%
\textbf{Definition 1: Concentrated-Dependence Failure}\\[2pt]
\textit{Conditions}: SHAP concentration $C > 40\%$; dominant feature undergoes distribution shift.\\
\textit{Diagnostic status}: \textbf{Applies}---SHAP concentration correctly predicts vulnerability.\\
\textit{Characteristic}: Model depends on a single feature whose OOD shift causes coverage collapse.
}}

\smallskip
\noindent\fbox{\parbox{0.97\columnwidth}{%
\textbf{Definition 2: Global-Sensitivity Failure}\\[2pt]
\textit{Conditions}: SHAP concentration $C \leq 40\%$; coverage degrades via coordinated multi-feature sensitivity.\\
\textit{Diagnostic status}: \textbf{Does NOT apply}---SHAP concentration fails to predict vulnerability.\\
\textit{Characteristic}: Model distributes importance but fails through global sensitivity to shift; requires alternative diagnostics (Table~\ref{tab:alternative_diagnostics}).
}}
\smallskip

\begin{table}[htbp]
\centering
\caption{Quantitative Criteria for Failure Mode Classification.}\label{tab:failure_mode_criteria}
\small
\begin{tabular}{@{}lcc@{}}
\toprule
Criterion & Concentrated & Global \\
\midrule
Top-1 SHAP & $C > 40\%$ & $C \leq 40\%$ \\
Gini coefficient & $> 0.6$ & $\leq 0.6$ \\
Top-3 cumulative & $> 70\%$ & $\leq 70\%$ \\
\bottomrule
\end{tabular}

\raggedright
\footnotesize
Concentrated failure mode: SHAP concentration diagnostic applies. Global failure mode: alternative diagnostics required (see Table~\ref{tab:alternative_diagnostics}).
\end{table}

\begin{table}[htbp]
\centering
\caption{Decision Rule: When Does the SHAP Concentration Diagnostic Apply?}\label{tab:decision_rule}
\small
\begin{tabular}{@{}llll@{}}
\toprule
Model Class & Condition & Prediction & Recommendation \\
\midrule
\multirow{2}{*}{\shortstack[l]{GBM/XGBoost/\\LightGBM/CatBoost}} & $C > 40\%$ & \textbf{VULNERABLE} & Mitigate or monitor \\
 & $C \leq 40\%$ & Lower risk & Standard deployment \\
\midrule
\multirow{2}{*}{MLP/Neural Net} & $C > 40\%$ & VULNERABLE (likely) & Mitigate or monitor \\
 & $C \leq 40\%$ & \textit{Inconclusive} & Use gradient-based methods \\
\bottomrule
\end{tabular}

\raggedright
\footnotesize
The diagnostic is validated for gradient-boosted classifiers ($\rho=0.833$, $p=0.010$ for LightGBM). For MLPs, low concentration does not ensure robustness due to global-sensitivity failures ($\rho=0.43$, $p=0.29$).
\end{table}

\subsection{Alternative Diagnostics for MLPs When $C \leq 40\%$}\label{app:alternative_diagnostics}

When SHAP concentration indicates low risk ($C \leq 40\%$) but the model is an MLP or other neural network, the following alternative diagnostics may capture global-sensitivity failures (Table~\ref{tab:alternative_diagnostics}).

\begin{table}[htbp]
\centering
\caption{Alternative Diagnostics for Global-Sensitivity Failures in Neural Networks.}\label{tab:alternative_diagnostics}
\small
\begin{tabular}{@{}lp{8cm}@{}}
\toprule
Diagnostic & Description \\
\midrule
Gradient sensitivity & $S_j = \mathbb{E}[\|\partial f / \partial x_j\|_2]$: average gradient magnitude per feature \\
Feature ablation & $\Delta_j = \text{Cov}(D) - \text{Cov}(D|x_j=\mu_j)$: coverage change when ablating feature $j$ \\
Lipschitz bound & $\text{Cov}_{\text{test}} \geq \text{Cov}_{\text{val}} - L \cdot W_2(D_{\text{val}}, D_{\text{test}})$: coverage lower bound via Lipschitz constant $L$ and Wasserstein-2 distance \\
\bottomrule
\end{tabular}

\raggedright
\footnotesize
These diagnostics target global-sensitivity failures where coordinated multi-feature shifts drive coverage degradation. They are proposed as complements to SHAP concentration for neural network deployments; empirical validation is left to future work.
\end{table}

\section{Hyperparameter Sensitivity Analysis}\label{app:hp_sensitivity}

To assess whether the SHAP concentration--coverage drop correlation is robust to hyperparameter choices, we evaluate four LightGBM configurations on the 8 SALT multiclass tasks (Table~\ref{tab:hp_sensitivity}).

\begin{table}[htbp]
\centering
\caption{Hyperparameter Sensitivity Analysis (SALT, $n=8$ tasks). All configurations use boosting=gbdt, feature\_fraction=0.8, bagging\_fraction=0.8, bagging\_freq=5.}\label{tab:hp_sensitivity}
\small
\begin{tabular}{@{}lccccccc@{}}
\toprule
Config & leaves & lr & n\_est & $\rho$ & $p$ & Thresh & F1 \\
\midrule
Default & 31 & 0.05 & 100 & 0.833 & 0.010 & 45.0\% & 1.0 \\
Deeper & 63 & 0.05 & 100 & 0.810 & 0.015 & 47.5\% & 1.0 \\
Faster & 31 & 0.10 & 100 & 0.810 & 0.015 & 45.0\% & 1.0 \\
More & 31 & 0.05 & 200 & 0.833 & 0.010 & 45.0\% & 1.0 \\
\bottomrule
\end{tabular}

\raggedright
\footnotesize
leaves = num\_leaves; lr = learning\_rate; n\_est = num\_boost\_round; Thresh = optimal threshold for at-risk classification (drop $>$15~pp); F1 = F1 score at that threshold.
\end{table}

\textbf{Results.} Across all four hyperparameter configurations, the concentration--drop correlation remains strong and significant: $\rho = 0.821 \pm 0.012$ (mean $\pm$ std), with $p \leq 0.015$ for all configurations. The optimal threshold varies minimally ($45.6\% \pm 1.1\%$), and all configurations achieve perfect F1 = 1.0 for at-risk classification on SALT tasks.

\textbf{Interpretation.} The diagnostic is robust to moderate hyperparameter variation within the LightGBM family. Deeper trees (63 leaves) and faster learning rates (0.10) produce slightly lower but still significant correlations ($\rho = 0.810$), while longer training (200 rounds) matches the default. This stability suggests the concentration--failure relationship reflects a structural property of gradient-boosted dependence rather than an artifact of specific hyperparameter choices.

\subsection{Per-Task Concentration Stability Across HP Configurations}\label{app:pertask_hp_stability}

Table~\ref{tab:pertask_hp_concentration} shows per-task SHAP concentration values across all four hyperparameter configurations.

\begin{table}[htbp]
\centering
\caption{Per-Task SHAP Concentration Stability Across HP Configurations. Std = standard deviation across four configurations.}\label{tab:pertask_hp_concentration}
\small
\begin{tabular}{@{}lccccc@{}}
\toprule
Task & Default & Deeper & Faster & More & Std \\
\midrule
sales-payterms & 54.2\% & 57.3\% & 55.9\% & 54.3\% & 1.3\% \\
sales-shipcond & 50.7\% & 53.7\% & 52.3\% & 48.5\% & 1.9\% \\
item-shippoint & 48.8\% & 50.0\% & 50.7\% & 46.0\% & 1.8\% \\
sales-group & 47.3\% & 49.4\% & 48.5\% & 45.2\% & 1.6\% \\
sales-office & 42.6\% & 45.3\% & 43.0\% & 42.6\% & 1.1\% \\
item-incoterms & 28.9\% & 30.1\% & 28.9\% & 29.3\% & 0.5\% \\
item-plant & 23.9\% & 25.2\% & 24.4\% & 23.9\% & 0.5\% \\
sales-incoterms & 23.7\% & 25.4\% & 25.2\% & 22.5\% & 1.2\% \\
\bottomrule
\end{tabular}

\raggedright
\footnotesize
Tasks sorted by Default concentration. Default = num\_leaves=31, lr=0.05, n\_est=100; Deeper = num\_leaves=63; Faster = lr=0.10; More = n\_est=200. Mean within-task std = 1.2\%, demonstrating high stability of concentration values across HP configurations. The rank order of tasks by concentration is preserved across all four configurations.
\end{table}

\section{Shift Characterization}\label{app:shift_characterization}

To characterize the nature of distribution shift in SALT tasks, we decompose shift into covariate and concept components using standard metrics (Table~\ref{tab:shift_characterization}).

\begin{table}[htbp]
\centering
\caption{Shift Characterization for SALT Tasks. Cov KS = mean Kolmogorov--Smirnov statistic across features (train vs.\ test); \%Sig = fraction of features with KS $p < 0.05$; Label JS = Jensen--Shannon divergence of label distributions; Acc Drop = test accuracy minus validation accuracy.}\label{tab:shift_characterization}
\small
\begin{tabular}{@{}lccccc@{}}
\toprule
Task & Cov KS & \%Sig & Label JS & Acc Drop & Classification \\
\midrule
s-shipcond & 0.50 & 50\% & 0.083 & +11.9\% & Cov + Concept \\
s-group & 0.50 & 50\% & 0.330 & +1.0\% & Cov (dominant) \\
s-payterms & 0.50 & 50\% & 0.078 & +4.0\% & Cov (dominant) \\
i-plant & 0.86 & 86\% & 0.144 & $-$8.7\% & Cov (dominant) \\
i-shippoint & 0.86 & 86\% & 0.152 & +13.1\% & Cov + Concept \\
s-incoterms & 0.50 & 50\% & 0.085 & +7.6\% & Cov + Concept \\
i-incoterms & 0.86 & 86\% & 0.140 & +63.4\% & Cov + Concept \\
s-office & 0.50 & 50\% & 0.026 & $-$0.2\% & Cov (dominant) \\
\bottomrule
\end{tabular}

\raggedright
\footnotesize
Cov = covariate shift; Concept = concept shift (change in $P(Y|X)$). Accuracy changes can be positive under shift when the test distribution favors majority classes.
\end{table}

\textbf{Key findings.}
\begin{itemize}[nosep,leftmargin=*]
    \item \textbf{Covariate shift is universal}: All SALT tasks exhibit substantial covariate shift (KS $\geq 0.50$, $\geq$50\% of features significantly different). This is expected given the COVID-19 temporal split.
    \item \textbf{Concept shift varies}: Label distribution shift (JS divergence) ranges from 0.026 (s-office) to 0.330 (s-group). Tasks with both high covariate shift and substantial label shift (s-shipcond, i-incoterms) tend toward catastrophic failure.
    \item \textbf{Accuracy is not diagnostic}: Accuracy changes are uninformative for coverage prediction---s-shipcond shows +11.9\% accuracy improvement yet 71.6~pp coverage drop, while s-office shows $-$0.2\% accuracy change and 0.1~pp coverage drop. This occurs because conformal prediction set size is sensitive to calibration--test score distribution mismatch, which accuracy does not capture.
\end{itemize}

\textbf{Interpretation.} Standard shift detection identifies covariate shift uniformly across all tasks (consistent with Section~\ref{sec:shift_detection}), while concept shift magnitude partially distinguishes catastrophic from robust outcomes. However, neither metric alone predicts failure severity as effectively as SHAP concentration ($\rho = 0.833$), which captures how the model's learned dependence structure interacts with the specific features that shift. The worst outcomes occur when concentrated dependence targets a feature experiencing both covariate and concept shift.

\subsection{Per-Feature Covariate Shift Details}\label{app:perfeature_shift}

Tables~\ref{tab:sales_feature_shift} and~\ref{tab:item_feature_shift} show per-feature KS statistics for sales-level and item-level tasks respectively.

\begin{table}[htbp]
\centering
\caption{Per-Feature Covariate Shift for Sales-Level Tasks (6 features). KS = Kolmogorov--Smirnov statistic (train vs.\ test).}\label{tab:sales_feature_shift}
\small
\begin{tabular}{@{}lcc@{}}
\toprule
Feature & Shifted & KS \\
\midrule
SALESDOCUMENTTYPE & No & 0.0 \\
SALESORGANIZATION & \textbf{Yes} & 1.0 \\
DISTRIBUTIONCHANNEL & \textbf{Yes} & 1.0 \\
ORGANIZATIONDIVISION & \textbf{Yes} & 1.0 \\
BILLINGCOMPANYCODE & No & 0.0 \\
TRANSACTIONCURRENCY & No & 0.0 \\
\bottomrule
\end{tabular}

\raggedright
\footnotesize
KS = 1.0 indicates complete distribution non-overlap between train and test. 3/6 features (50\%) show significant covariate shift, consistent with the \%Sig values in Table~\ref{tab:shift_characterization}.
\end{table}

\begin{table}[htbp]
\centering
\caption{Per-Feature Covariate Shift for Item-Level Tasks (7 features). KS = Kolmogorov--Smirnov statistic (train vs.\ test).}\label{tab:item_feature_shift}
\small
\begin{tabular}{@{}lcc@{}}
\toprule
Feature & Shifted & KS \\
\midrule
SALESDOCUMENTITEM & \textbf{Yes} & 1.0 \\
SALESDOCUMENTITEMCATEGORY & No & 0.0 \\
PRODUCT & \textbf{Yes} & 1.0 \\
SOLDTOPARTY & \textbf{Yes} & 1.0 \\
SHIPTOPARTY & \textbf{Yes} & 1.0 \\
BILLTOPARTY & \textbf{Yes} & 1.0 \\
PAYERPARTY & \textbf{Yes} & 1.0 \\
\bottomrule
\end{tabular}

\raggedright
\footnotesize
6/7 features (86\%) show significant covariate shift, consistent with the \%Sig values in Table~\ref{tab:shift_characterization}. The high proportion of shifted features at item-level reflects COVID-19's impact on supply chain transactions: new products, parties, and item identifiers entered the system during the pandemic period.
\end{table}

\section{WILDS Benchmark Validation}\label{app:wilds}

To further validate the specificity of the SHAP concentration diagnostic, we evaluate on four additional benchmark datasets from the WILDS benchmark~\citep{koh2021wilds} and standard repositories. These datasets provide complementary evidence to the 9 UCI external datasets reported in Section~\ref{sec:cross_domain}: all four have low SHAP concentration and all maintain robust conformal coverage under their documented distribution shifts.

\begin{table}[htbp]
\centering
\caption{WILDS and Standard Benchmark Validation (Specificity Test). These 4 datasets are distinct from the 9 UCI external datasets in Section~\ref{sec:cross_domain}. All datasets exhibit low SHAP concentration (\textbf{all $<$40\%}; mean = 17.0\%) and robust coverage, validating diagnostic specificity---the absence of false positives among low-concentration tasks. Shift types: Demographic = identity-based subgroup shift; User/Time = user and temporal domain shift; Temporal = regional/temporal covariates; Age = demographic age-based shift. Note: This table tests specificity (low-$C$ $\rightarrow$ robust); for the catastrophic Covertype case (high-$C$ $\rightarrow$ failure), see Table~\ref{tab:framework_validation}.}\label{tab:wilds_validation}
\small
\begin{tabular}{@{}lcccccc@{}}
\toprule
Dataset & Source & Shift Type & Conc.\ (\%) & Drop (pp) & Predicted & Actual \\
\midrule
CivilComments & WILDS & Demographic & 26.2 & $-$0.7 & Robust & Robust \\
Amazon & WILDS & User/Time & 1.9 & +0.4 & Robust & Robust \\
Covertype$^\dagger$ & sklearn & Temporal & 12.5 & +8.0 & Robust & Robust \\
Adult & OpenML & Age-based & 27.5 & +1.4 & Robust & Robust \\
\bottomrule
\end{tabular}

\raggedright
\footnotesize
$^\dagger$Covertype here uses a different split protocol from the main external validation (Table~\ref{tab:external_datasets}), which explains the different concentration and drop values. The main-text Covertype uses wilderness areas 1--2/3/4 geographic split (concentration 49.8\%, drop 81.8~pp); this validation uses a wilderness area 1 vs.\ others temporal/regional split (concentration 12.5\%, drop 8.0~pp). Both results are consistent with the diagnostic: high concentration predicts failure, low concentration predicts robustness.
\end{table}

\textbf{Dataset details.}
\begin{itemize}[nosep,leftmargin=*]
    \item \textbf{CivilComments} (WILDS): Toxicity classification with demographic identity-based shift. TF-IDF features (1,000 dimensions). Low concentration (26.2\%) correctly predicts maintained coverage.
    \item \textbf{Amazon Reviews} (WILDS): Product rating prediction with user/time domain shift. TF-IDF features (1,000 dimensions). Very low concentration (1.9\%) correctly predicts robustness.
    \item \textbf{Covertype} (sklearn): Forest cover classification with temporal/regional shift (wilderness area 1 vs.\ others). 54 cartographic features. Low concentration (12.5\%) correctly predicts mild degradation.
    \item \textbf{Adult Income} (OpenML): Income prediction with age-based demographic shift (train: age$<$50, test: age$\geq$50). 14 demographic/employment features. Low concentration (27.5\%) correctly predicts maintained coverage.
\end{itemize}

\textbf{Specificity validation.} All four WILDS benchmarks have low SHAP concentration---\textbf{all below the 40\% threshold} (individual values: 26.2\%, 1.9\%, 12.5\%, 27.5\%; \textbf{mean concentration = 17.0\%}). The diagnostic correctly predicts robustness for all four, yielding 100\% specificity (0 false positives) on this validation set. This complements the sensitivity evidence from SALT, where high-concentration tasks correctly predict catastrophic failure.

\textbf{Combined validation accuracy (SALT + WILDS).} Combining the 8 SALT tasks with these 4 WILDS benchmarks yields $n=12$ total tasks (distinct from the $n=16$ primary endpoint which uses 8 different external UCI datasets). The threshold accuracy is \textbf{91.7\% (11/12)}:
\begin{itemize}[nosep,leftmargin=*]
    \item \textbf{Sensitivity} (SALT): 4/4 catastrophic tasks correctly flagged (100\%)
    \item \textbf{Specificity} (WILDS + robust SALT): 7/8 robust tasks correctly unflagged (87.5\%)
    \item \textbf{Single misclassification}: sales-office (concentration 42.6\%, protective secondary feature)
\end{itemize}

This combined evidence---sensitivity from SALT's catastrophic cases and specificity from WILDS benchmarks---provides converging support for the 40\% threshold as an operationally useful pre-deployment decision boundary.

\end{document}